\documentclass[journal]{IEEEtran}

\usepackage[utf8]{inputenc} 
\usepackage[T1]{fontenc}    
\usepackage{url}            
\usepackage{cite}
\usepackage{booktabs}       
\usepackage{amsfonts}       
\usepackage{nicefrac}       
\usepackage{microtype}      
\usepackage{threeparttable}
\usepackage{makeidx}
\usepackage{indentfirst}
\usepackage{amssymb}
\usepackage{mathtools}
\usepackage{amsmath}
\usepackage{algorithm}
\usepackage{algorithmic}
\usepackage{subcaption}
\usepackage{graphicx}
\usepackage{multirow}
\usepackage{adjustbox}
\usepackage{makecell}
\usepackage{hhline}
\usepackage{xr}
\usepackage{xcolor}
\usepackage{lineno, hyperref}

\begin{document}
\title{Self-supervised Representation Learning for Evolutionary Neural Architecture Search}

\author{Chen Wei, Yiping Tang, Chuang Niu, Haihong Hu, Yue Wang and
Jimin Liang,~\IEEEmembership{Member,~IEEE,}
\thanks{Manuscript received xxxx xx, 2020; revised xxxx xx, 2020; accepted xxxx xx, 2020. This work was supported in part by the National Natural Science Foundation of China under Grants U19B2030 and 61976167, and in part by the Xi'an Science and Technology Program under Grant 201809170CX11JC12. {\sl (Corresponding author: Jimin Liang.)}}
\thanks{Chen Wei is with the School of Electronic Engineering, Xidian University, Xi’an, Shaanxi 710071, China, and also with the College of Economics and Management, Xi'an University of Posts\&Telecommunications, Xi'an, Shaanxi 710061, China (e-mail: weichen\_3@stu.xidian.edu.cn).}
\thanks{Yiping Tang, Haihong Hu, Yue Wang and Jimin Liang are with School of Electronic Engineering, Xidian University, Xi’an, Shaanxi 710071, China (e-mail: tangyiping@aliyun.com; hhhu@mail.xidian.edu.cn; wangyue1991@stu.xidian.edu.cn; jimleung@mail.xidian.edu.cn).}
\thanks{Chuang Niu is with the Department of Biomedical Engineering, Rensselaer Polytechnic Institute, Troy, NY 12180, USA (e-mail: niuc@rpi.edu)}
}


\IEEEtitleabstractindextext{%
  \begin{abstract}
    Recently proposed neural architecture search (NAS) algorithms adopt neural predictors to accelerate the architecture search. The capability of neural predictors to accurately predict the performance metrics of neural architecture is critical to NAS, and the acquisition of training datasets for neural predictors is time-consuming. How to obtain a neural predictor with high prediction accuracy using a small amount of training data is a central problem to neural predictor-based NAS. Here, we firstly design a new architecture encoding scheme that overcomes the drawbacks of existing vector-based architecture encoding schemes to calculate the graph edit distance of neural architectures. To enhance the predictive performance of neural predictors, we devise two self-supervised learning methods from different perspectives to pre-train the architecture embedding part of neural predictors to generate a meaningful representation of neural architectures. The first one is to train a carefully designed two branch graph neural network model to predict the graph edit distance of two input neural architectures. The second method is inspired by the prevalently contrastive learning, and we present a new contrastive learning algorithm that utilizes a central feature vector as a proxy to contrast positive pairs against negative pairs. Experimental results illustrate that the pre-trained neural predictors can achieve comparable or superior performance compared with their supervised counterparts with several times less training samples. We achieve state-of-the-art performance on the NASBench-101 and NASBench201 benchmarks when integrating the pre-trained neural predictors with an evolutionary NAS algorithm.
  
  \end{abstract}
  
  }

\maketitle

\IEEEdisplaynontitleabstractindextext

\IEEEpeerreviewmaketitle

\section{Introduction}
\IEEEPARstart{N}{eural} architecture search (NAS) refers to the use of certain search strategies to find the best performing neural architecture in a pre-defined search space with minimum searching costs \cite{Elsken2018NeuralAS}. The search strategies sample the potentially promising neural architectures from the search space and the performance metrics of the sampled architectures, obtained from time-consuming training and validation procedures, are used to optimize the search strategies. To alleviate the time cost of training and validation procedures, some recently proposed NAS search strategies employ neural predictors to accelerate the performance estimation of sampled architectures \cite{Liu2017ProgressiveNA,White2019BANANASBO,Wang2019AlphaXEN,DBLP:journals/corr/abs-2004-01899,Chen2019NPENAS}. The capability of neural predictors to accurately predict the performance of sampled architectures is critical to downstream search strategies \cite{Liu2017ProgressiveNA, DBLP:journals/corr/abs-2004-01899, Chen2019NPENAS, DBLP:conf/eccv/WenLCLBK20, DBLP:journals/corr/abs-2006-06936}.Because of the significant time cost of obtaining labeled training samples, how to acquire accurate neural predictors using a fewer number of training samples is one of the key issues in NAS methods employing neural predictors.

Self-supervised representation learning, a type of unsupervised representation learning, has been successfully applied in areas such as image classification \cite{DBLP:journals/corr/abs-2002-05709,DBLP:conf/cvpr/He0WXG20} and natural language processing \cite{DBLP:conf/naacl/DevlinCLT19}. If a model is pre-trained by an effective self-supervised representation learning and then fine-tuned by supervised learning using a few labeled training data, then it is highly likely to outperform its supervised counterparts \cite{DBLP:journals/corr/abs-2002-05709, DBLP:conf/cvpr/He0WXG20, DBLP:conf/iclr/AsanoRV20a}. In this paper, we novelty study and apply the self-supervised representation learning to the NAS domain to enhance the performance of neural predictors built from graph neural networks \cite{Wu2020ACS} and employ it to the downstream evolutionary search strategy.

Effective unsupervised representation learning falls into one of two categories: generative or discriminative \cite{DBLP:journals/corr/abs-2002-05709}. Existing unsupervised representation learning methods for NAS \cite{DBLP:journals/corr/abs-2006-06936, DBLP:journals/corr/abs-2007-04452} belong to the generative category. Their learning objective is to make the neural predictor correctly reconstruct the input neural architecture, but it has limited relevance to NAS. This may result in the trained neural predictor producing a less effective representation of the input neural architecture. Discriminative unsupervised representation learning, also known as self-supervised learning, requires designing a pretext task \cite{DBLP:conf/eccv/NorooziF16, DBLP:conf/iclr/GidarisSK18} from an unlabeled dataset and using it as supervision to learn meaningful feature representation. Inspired by previous findings that ``close by'' architectures tend to have similar performance metrics \cite{Ying2019NASBench101TR, DBLP:journals/corr/abs-2007-06559}, we adopt the graph edit distance (GED) as a supervision to carry out self-supervised learning because GED can reflect the distance of different neural architectures in the search space. Commonly used GED is computed based on the graph encoding of two different neural architectures (adjacency matrices and node operations) \cite{Ying2019NASBench101TR}, but this scheme cannot identify isomorphic graphs. Path-based encoding \cite{White2019BANANASBO} is another commonly used neural architecture encoding scheme, but it can not recognize the position of each operation in the neural architecture, e.g., two different operations in a neural architecture may have the same path-encoding vectors. To overcome the above drawbacks, we propose a new neural architecture encoding scheme denoted as position-aware path-based encoding, which can identify isomorphic graphs and recognize the position of different operations in neural architectures.

Since different pretext tasks may lead to different feature representations, utilizing the GED, we devise two self-supervised learning methods from two different perspectives to improve the feature representation of neural architectures, and to investigate the effect of different pretext tasks on the predictive performance of neural predictors. The first method utilizes a handcrafted pretext task, while the second one learns feature representation by contrasting positive pairs against negative pairs. 


The pretext task of the first self-supervised learning method is to predict the normalized GED of two different neural architectures in the search space. We design a model with two independent identical branches and use the concatenation of their output features to predict the normalized GED. After the self-supervised pre-training, we adopt only one branch of the model to build neural predictor. This method is denoted as self-supervised regression learning. 

The second self-supervised learning method is inspired by the prevalently contrastive learning for image classification \cite{DBLP:journals/corr/abs-2002-05709,DBLP:conf/cvpr/He0WXG20,DBLP:conf/iclr/AsanoRV20a}, which maximizes the agreement between differently augmented views of the same image via a contrastive loss in the latent space \cite{DBLP:journals/corr/abs-2002-05709}. Since there is no guarantee that a neural architecture and its transformed form will have the same performance metrics, it is not reasonable to directly apply the contrastive learning to NAS. We propose a new contrastive learning algorithm, termed central contrastive learning, that uses the feature vector of a neural architecture and its nearby neural architectures' feature vectors (with small GEDs) to build a central feature vector. Then the contrastive loss is utilized to tightly aggregate the feature vectors of the architecture and its nearby architectures onto the central feature vector and push the feature vectors of other neural architectures away from the central feature vector. This method is indicated as self-supervised central contrastive learning. 


After self-supervised pre-training, two neural predictors are built by connecting a fully connected layer to the architecture embedding modules of the pre-trained models. Finally, we integrate the pre-trained neural predictors into the neural predictor guided evolution neural architecture search (NPENAS) algorithm \cite{Chen2019NPENAS} to verify their performance. 

Our main contributions can be summarized as follows.
\begin{itemize}
  \item We present a new vector encoding scheme, termed position-aware path-based encoding, to overcome the drawbacks of the adjacency matrix encoding and path-based encoding methods. The scheme is more efficient than path-based encoding, and the experimental results illustrate its superiority to filter out isomorphic graphs.
  \item We propose a self-supervised regression learning method that defines a pretext task to predict the normalized GED of two different neural architectures and design a neural network with two independent identical branches to learn meaningful representation of neural architectures. After the self-supervised pre-training, a neural predictor is constructed and fine-tuned. The neural predictor pre-trained by this method achieves its performance upper bound with a small search budget and a few training epochs, and in the best case, achieves better performance using ten times less search budget than it supervised counterparts.
  
  \item We present a central contrastive learning algorithm that forces neural architectures with small GED to lean closer together in the feature space, while neural architectures with large GED divide further apart. When trained for more epochs, the pre-trained neural predictor fine-tuned with half search budget can achieve comparable performance to its supervised counterparts, and with the same search budget, the fine-tuned neural predictor outperforms their supervised counterparts by about 1.5 times. The proposed central contrastive learning algorithm can also be extended to the domain of graph unsupervised representation learning without any modifications.
  \item When integrating the pre-trained neural predictors to the NPENAS, we achieve state-of-the-art performance on the NASBench-101 \cite{Ying2019NASBench101TR} and NASBench-201 \cite{Dong2020NASBench201ET} benchmarks. The searched neural architectures have comparable or equal results to the \textit{ORACLE} baseline (performance upper bound). 
\end{itemize}
\section{Related Works}
\subsection{Neural Architecture Search} \label{nas}
Due to the huge size of the pre-defined search space, NAS usually search for the potential superiority neural network architectures by utilizing a search strategy. Reinforcement learning (RL) \cite{zoph2016neural, enas, Zoph2017LearningTA, Liu2017ProgressiveNA}, evolutionary algorithms \cite{Real2017LargeScaleEO, Real2018RegularizedEF, 9075201, 8742788, 8712430, Chen2019NPENAS}, gradient-based methods \cite{Liu2018DARTSDA, Zhou2019BayesNASAB, Xie2019SNASSN, Chen2019ProgressiveDA, xu2020pcdarts}, Bayesian optimization (BO) \cite{kandasamy2018neural,White2019BANANASBO,Chen2019NPENAS}, and predictor-based methods \cite{DBLP:conf/eccv/WenLCLBK20, Wang2019AlphaXEN, DBLP:journals/corr/abs-2004-01899, DBLP:journals/corr/abs-2006-06936} are the commonly used search strategies. A search strategy adjusts itself by exploiting the selected neural architectures' performance metrics to explore the search space better.

As it is time-consuming to estimate the performance metrics of a given neural architecture through the training and validation procedures, many performance estimation strategies are proposed to speed up this task. Commonly used strategies include using a proxy dataset and proxy architecture, early stopping, inheriting weights from a trained architecture, and weight sharing \cite{Elsken2018NeuralAS}. A neural predictor that is employed to estimate the performance metrics of the neural network architectures can also be recognized as a kind of performance estimation strategy. Recently, many NAS algorithms have adopted neural predictors to explore the search space \cite{Liu2017ProgressiveNA, White2019BANANASBO, DBLP:conf/eccv/WenLCLBK20, Wang2019AlphaXEN, DBLP:journals/corr/abs-2004-01899, Chen2019NPENAS}. The capability of neural predictors to accurately predict the performance of neural architectures is critical for NAS algorithms using neural predictors. The neural predictors are trained on a training dataset that is composed by some neural architectures together with their corresponding performance metrics, which are acquired through the time-consuming training and validation procedures. 

In this paper, we novelty apply the self-supervised representation learning to the NAS domain. We propose two self-supervised representation learning methods to improve the feature representation of neural predictors that are built from graph neural networks, thus enhance the prediction performance of neural predictors.


\subsection{Neural Architecture Encoding Scheme} \label{nae}
The neural architecture is usually defined as a direct acyclic graph (DAG). The adjacency matrix of the graph is used to represent the connections of operations, and the nodes are used to represent the operations. The commonly used neural architecture encoding schemes can be categorized into vector encoding scheme and graph encoding scheme. 

Adjacency matrix encoding \cite{Wang2019AlphaXEN, Zhou2019BayesNASAB, DBLP:conf/iclr/BakerGRN18} and path-based encoding \cite{White2019BANANASBO} are two frequently used vector encoding schemes. The adjacency matrix encoding is the concatenation of the flattened adjacency matrix and the one-hot encoding vector of each node, but it can not identify the isomorphic graphs \cite{DBLP:journals/corr/abs-2006-06936}. The path-based encoding is the encoding of the input-to-output paths of the neural architecture, but as demonstrated in Appendix \ref{appendices_1}, this scheme can not recognize the position of operations in the neural architecture. The graph encoding scheme represents the neural architecture by its adjacency matrix and the one-hot encoding of each node. 

In this paper, we propose a new vector encoding scheme denoted as position-aware path-based encoding. This encoding scheme can identify the isomorphic graphs and recognize the position of operations in the neural architecture. We adopt the graph encoding scheme and employ a graph neural network \cite{Wu2020ACS} to embed the neural architecture into feature space. Since the graph encoding scheme can not identify isomorphic graphs \cite{DBLP:journals/corr/abs-2006-06936}, the position-aware path-based encoding is used firstly to filter out isomorphic graphs. We also utilize the position-aware path-based encoding to calculate the GED of different neural architectures. 

\subsection{Unsupervised Representation Learning for NAS} \label{url}
Unsupervised representation learning methods fall into to two categories - generative and discriminative \cite{DBLP:journals/corr/abs-2002-05709}. 
The learning objective of existing generative unsupervised learning methods for NAS, arc2vec \cite{DBLP:journals/corr/abs-2006-06936} and NASGEM \cite{DBLP:journals/corr/abs-2007-04452}, is to reconstruct the input neural architectures using an encoder-decoder network, which has little relevance to NAS. Moreover, the arc2vec \cite{DBLP:journals/corr/abs-2006-06936} adopts the variational autoencoder \cite{DBLP:journals/corr/KingmaW13} to embed the input neural architectures into a high dimensional continuous feature space, and the feature space is assumed to follow the Gaussian distribution. Since there is no guarantee that the real underlying distribution of the feature space is Gaussian, this assumption may harm the representation of neural architectures. The NASGEM \cite{DBLP:journals/corr/abs-2007-04452} adds a similarity loss to improve the feature representation. However, the similarity loss only considers the adjacency matrix of the input neural architecture and ignores the node operations, resulting in the failure to identify isomorphic graphs.

We present two self-supervised learning methods to the domain of NAS. The first one is inspired by unsupervised graph representation learning. GMNs \cite{DBLP:conf/icml/LiGDVK19} adopts graph neural network as building blocks and presents a cross-graph attention-based mechanism to predict the similarity of the two input graphs. SimGNN \cite{DBLP:conf/wsdm/BaiDBCSW19} takes two graphs as input, embeds the graph and each node of the graph into the feature space using a graph convolutional neural network, and then uses graph feature similarity and node feature similarity to predict the similarity of the input graphs. UGRAPHEMB \cite{DBLP:conf/ijcai/BaiDQMG0SW19} takes two graphs as input, adopts graph isomorphism network (GIN) \cite{Xu2018HowPA} to embed the input graphs into feature space, and utilizes a multi-scale node attention mechanism to predict the similarity of the input graphs. Our work is quite like UGRAPHEMB, but we design a new neural network model without using the complex multi-scale node attention and apply the unsupervised learning to the field of neural architecture representation learning.


The second method is inspired by the contrastive learning for image classification. The contrastive learning used in image classification forces the image and its transformations to be similar in the feature space \cite{DBLP:journals/corr/abs-2002-05709, DBLP:conf/cvpr/He0WXG20, DBLP:journals/corr/abs-2006-09882}. Since there is no guarantee that a neural architecture and its transformed form will have the same performance metrics, it is not reasonable to directly apply the contrastive learning for image classification to the NAS domain. We propose a new contrastive learning algorithm, central contrast learning, to learn meaningful representation of neural architectures. To our best knowledge, this is the first study applying contrastive learning to the NAS domain and the field of unsupervised graph similarity learning.

\section{Methodology}
To enhance the prediction performance of neural predictors, we propose two self-supervised representation learning methods to improve the feature representation ability of the neural predictors. We design a new neural architecture encoding scheme to calculate the GED of graphs in Section \ref{papbe}. The self-supervised regression learning that utilizes a carefully designed model with two independent identical graph neural network branches to predict the GED of neural architectures is discussed in Section \ref{snnssl}. The self-supervised central contrastive learning is introduced in Section \ref{ccl}. The utilization of the pre-trained neural predictors for the downstream search strategies is elaborated in Section \ref{nas}.

\begin{figure*}[ht!]
  \centering
  \includegraphics[width=0.87\textwidth, height=7cm]{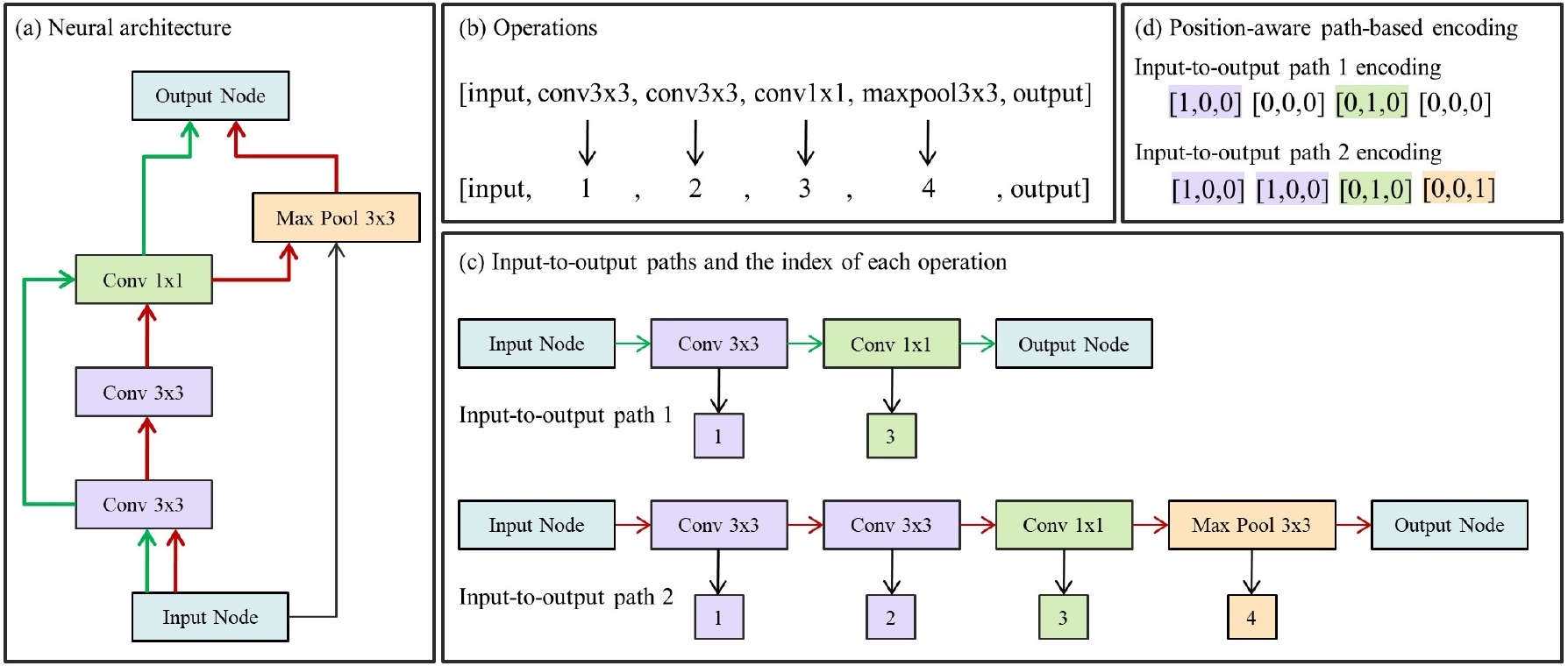}
  \caption{Overview of the position-aware path-based encoding. (a) A neural architecture in the NASBench-101 search space. The green and red lines indicate two input-to-output paths. (b) Operations and their corresponding unique indices. (c) Two different input-to-output paths and their operation indices. (d) Position-aware path-based encoding of the two input-to-output paths in (c).}
  \label{fig_1_papbe}
\end{figure*}

\subsection{Problem Formulation}
In a pre-defined search space $S$, a neural architecture $s$ can be represented as a DAG
\begin{equation} \label{graph_define}
  s=(V, E), \quad s \in S, 
\end{equation}
where $V=\{v_i\}_{i=1:H}$ is the set of nodes representing operations in $s$, $E=\{v_i, v_j\}_{i,j=1:H}$ is the set of edges describing the connection of operations, and $H$ is the number of nodes.

Predictor-based NAS adopts a neural predictor modeled as
\begin{equation}
  \hat y = f(s),
\end{equation}
where $f$ is the neural predictor, and it takes a neural architecture $s$ as input and outputs the performance metric prediction $\hat y$ of $s$.


\subsection{Position-aware Path-based Encoding} \label{papbe}


Since the proposed self-supervised learning methods utilize the GED to measure the similarity of different neural architectures, it is critical to calculate the GED effectively. We present a new vector encoding scheme, position-aware path-based encoding, which improves on the path-based encoding \cite{White2019BANANASBO} by recording the position of each operation in the path. The scheme consists of two steps: generating the position-aware path-based encoding vectors for the input-to-output paths of the neural architecture, and concatenating the vectors of all the paths.

As shown in Eq \ref{graph_define}, a neural architecture can be defined by a DAG with its nodes representing the operations in the neural architecture. The DAG consists of an input node, some operation nodes and an output node, connected in sequence. The adjacency matrix of the DAG is used to represent the connections of the different nodes. Since each node in the DAG has a fixed position, we assign each node with a unique index, which implies that each operation associated with the node has a unique index. 

NASBench-101 \cite{Ying2019NASBench101TR} is a widely used NAS search space. It contains three different operations: convolution 3$\times$3, convolution 1$\times$1, and max-pool 3$\times$3. Fig. \ref{fig_1_papbe}a illustrates a neural architecture in the NASBench-101 search space and uses green and red lines to indicate two different input-to-output paths. The operations and their corresponding indices of the neural architecture are shown in Fig. \ref{fig_1_papbe}b. In Fig. \ref{fig_1_papbe}c, we demonstrate two input-to-output paths of the neural architecture in Fig. \ref{fig_1_papbe}a.  Unlike the path-based encoding \cite{White2019BANANASBO}, when we extract all the input-to-output paths in the neural architecture, we also record the index of operations in the input-to-output paths. The position-aware path-based encoding of the two different paths in Fig. \ref{fig_1_papbe}c is indicated in Fig. \ref{fig_1_papbe}d. The vector length of each input-to-output path is fixed, which equals the multiplication of the number of operation nodes and the number of operation types. We traverse all the operation nodes in the neural architecture. If an operation node appears in the input-to-output path, then the operation is represented by its one-hot operation type vector; otherwise, it is represented by a zero vector. Since there are three different operations in the NASBench-101, the length of the one-hot operation vector and the zero vector is three.

The final encoding vector is the concatenation of the position-aware path-based encoding vectors for all the input-to-output paths in the neural architecture. To keep the concatenation consistent, we design the following steps:

\begin{enumerate}
  \item Firstly, sort all the input-to-output paths in ascending order by path length.
  \item Secondly, sort all the input-to-output paths of the same path length in ascending order by the operation index.
  \item Finally, concatenate the sorted input-to-output paths' position-aware path-based encoding vectors.
\end{enumerate}

The vector length of path-based encoding \cite{White2019BANANASBO} increases exponentially with the number of operation nodes, whereas the vector length of the position-aware path-based encoding increases linearly with the number of input-to-output paths. Therefore, the position-aware path-based encoding is a more efficient vector encoding scheme than path-based encoding. As the number of input-to-output paths may be different in different neural architectures, we will pad the short vectors with zeros to keep all vectors have the same length.

\subsection{Self-supervised Regression Learning} \label{snnssl}
The pretext task of the proposed self-supervised regression learning is to predict the normalized GED of two input neural architectures. The GED is defined as 
\begin{equation} \label{eq_ged}
  GED(s_i, s_j) = \sum_{k=i}^K \lvert \textbf{p}_i^k - \textbf{p}_j^k \rvert, \quad s_i, s_j \in S,
\end{equation}
where $\textbf{p}_i$ and $\textbf{p}_j$ are the position-aware path-based encoding vectors of architecture $s_i$ and $s_j$, $\textbf{p}_i^k$ and $\textbf{p}_j^k$ are the $k^{th}$ elements of the position-aware encoding vector $\textbf{p}_i$ and $\textbf{p}_j$, and $K$ is the vector length.

Following \cite{DBLP:conf/wsdm/BaiDBCSW19}, we define the normalized GED as 
\begin{equation}
    nGED(s_i, s_j) = exp^{-dist}, \quad where  \quad dist=\frac{GED(s_i, s_j)}{|V|},
\end{equation}
where $|V|$ is the number of nodes in the neural architectures.

As the architecture in search space is represented as a DAG, it is straightforward to adopt graph neural networks to aggregate features for each node and generate the graph embedding by averaging the nodes' features. We design self-supervised models and neural predictors utilizing the spatial-based graph neural network GIN layers.

\begin{figure}[ht!]
  \begin{center}
      \includegraphics[width=0.9\linewidth, height=3.7cm]{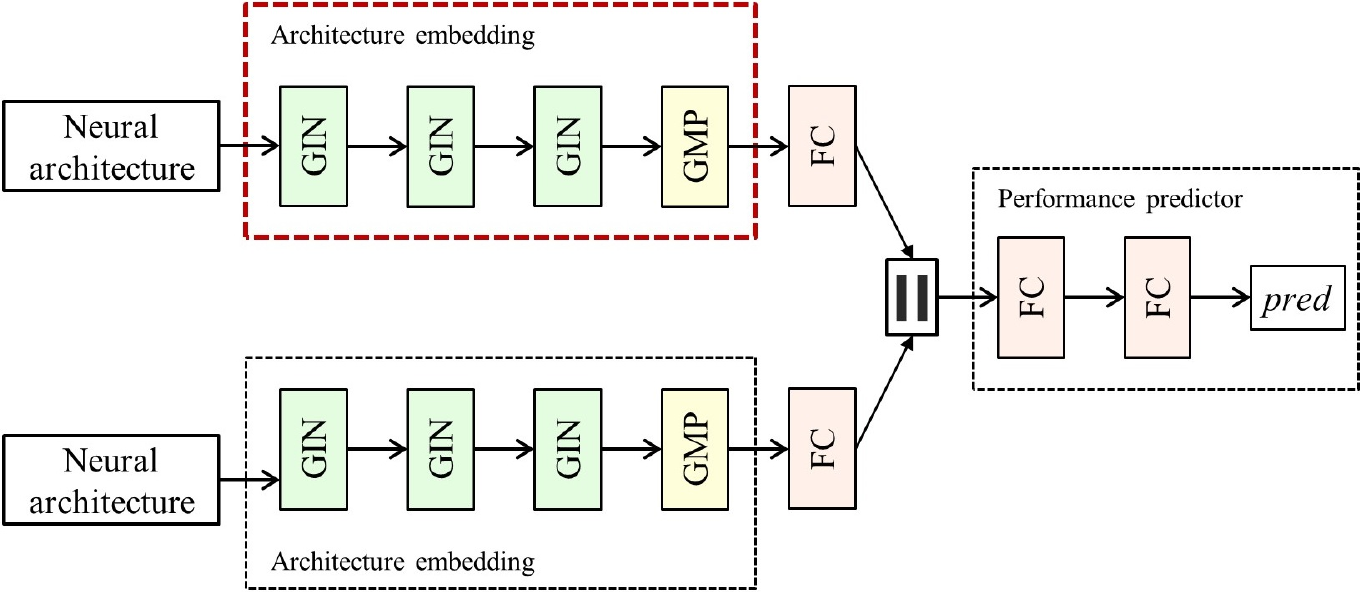}
      \caption{Structure of model $f_{rl}$.}
      \label{fig_2}
  \end{center}
\end{figure}

Since the pretext task is to predict the normalized GED of two different neural architectures, we design a regression model $f_{rl}$ that consists of two independent identical graph neural network branches, as illustrated in Fig. \ref{fig_2}. Each branch is composed of three sequentially connected GIN layers and a global mean pooling (GMP) layer. The GMP layer outputs the mean of node features of the last GIN layer. The outputs of the two branches are concatenated, and then sent to two sequentially connected fully connected layers to predict the two input architectures' normalized GED. The regression loss function to optimize the parameters $w_{rl}$ of $f_{rl}$ is formulated as 

\begin{equation}
  w_{rl}^* = \arg\min_{w_{rl}} \sum_{(s_i, s_j) \in S} (f_{rl}(s_i, s_j) - nGED(s_i, s_j))^2.
\end{equation}

After the self-supervised pre-training, we can select any branch of $f_{rl}$ to embed the neural architectures into feature space. We design a neural predictor by connecting a fully connected layer to the architecture embedding module (as illustrated in the red rectangle of Fig. \ref{fig_2}) of the pre-trained models. The regression loss is employed to fine-tune the neural predictor. The parameters of the neural predictor, denoted as $w$, are optimized as 
\begin{equation} \label{regression_loss_precitor}
  w^* = \arg\min_{w} \sum_{s_i \in S} (f(s_i) - y_i)^2, 
\end{equation} 
where $y_i$ is the performance metric of $s_i$.

\subsection{Self-supervised Central Contrastive Learning} \label{ccl}
We present a central contrastive learning algorithm to force neural architectures with small GED to lean closer together in the feature space, while neural architectures with large GED divide further apart. 

\begin{figure}[ht!]
  \begin{center}
      \includegraphics[width=0.94\linewidth, height=1.5cm]{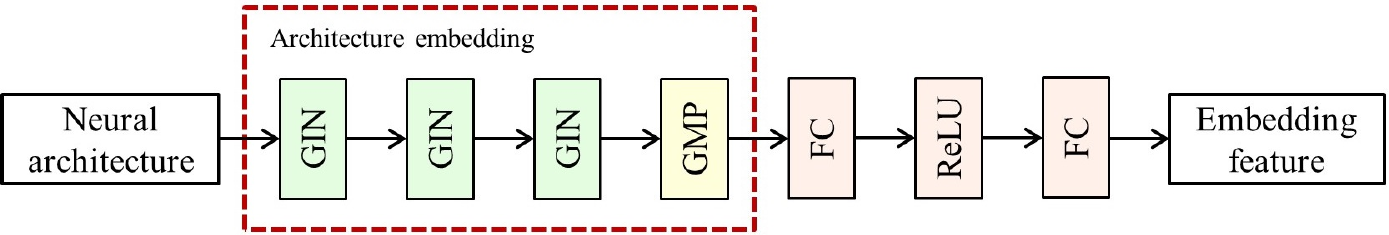}
      \caption{Structure of model $f_{ccl}$.}
      \label{fig_3}
  \end{center}
\end{figure}

 
As illustrated in Fig. \ref{fig_3}, we design a model $f_{ccl}$ to embed the neural architecture into feature space. Following the SimCLR's \cite{DBLP:journals/corr/abs-2002-05709}, $f_{ccl}$ consists of a neural architecture embedding module, a non-linear fully connected layer, and a fully connected layer. For a fair comparison, the architecture embedding module is identical to that of $f_{rl}$. After the self-supervised central contrastive pre-training, we connect a fully connect layer to the architecture embedding module to predict the input neural architecture's performance.

Given a batch of neural architectures $S_b = \{s_k\}_{k=1}^{N}$ and a neural architecture $ s_i \in S_b $, we first calculate the minimum GED of $s_i$ to all other architectures $s_j \in S_b$. Second, we collect the neural architectures $s_k \in S_b$ that have the same $GED(s_i, s_k)$ as the minimum GED, and denote the set of collections as $S_{pos}$. We also put $s_i$ into $S_{pos}$. The set of neural architectures in this batch but not in set $S_{pos}$ is denoted as $S_{neg}$. We then use the model $f_{ccl}$ to embed all the neural architectures in $S_{pos}$ and $S_{neg}$ into the feature space, and denote $E_{pos}$ as the feature vector set of $S_{pos}$ and $E_{neg}$ as the feature vector set of $S_{neg}$. A central vector $\textbf{e}_c$ is calculated by averaging all the feature vectors in $E_{pos}$. At last, the contrastive loss is used to aggregate all the feature vectors in $E_{pos}$ to the central vector $\textbf{e}_c$ and push the feature vectors in $E_{neg}$ far away from $\textbf{e}_c$. An example of the central contrastive learning is illustrated in Appendix \ref{appendices_2}.

The detailed procedure of central contrastive learning is summarized in the Algorithm. \ref{algo:1}, where the feature vectors $\textbf{e}_j$ and $\textbf{e}_c$ are normalized vectors.

\begin{algorithm}
  \caption{\label{algo:1} Central Contrast Learning}
  \begin{algorithmic}[1]
  \STATE \textbf{Input:} batch size $N$, number of training architectures $M$ and $M \le N$, temperature $\tau$, regularization weight $\lambda$, model $f_{ccl}$.
  \FOR{sampled minibatch $S_b = \{s_k\}_{k=1}^{N}$}
    \STATE $E_c = \varnothing $
    \STATE \textbf{for all} $t\in \{1, \ldots, M\}$ \textbf{do}
    \STATE $~~~~$ randomly draw one neural architecture $s_i \in S_b$
    \STATE $~~~~$ $g_{min} = \min GED(s_i, s_j)$ where $j=\{1, \ldots, i-1,i+1,\ldots, N\}$ \textcolor{gray}{\# $GED$: Eq \ref{eq_ged}} 
    \STATE $~~~~$ $E_{pos} = \varnothing$ and $E_{neg} = \varnothing $ 
    \STATE $~~~~$ \textbf{for all} $ j \in \{1, \ldots,i-1, i+1, \ldots N\}$ \textbf{do}
    \STATE $~~~~~~~~$ $\textbf{e}_j = f_{ccl}(s_j)$ 
    \STATE $~~~~~~~~$ \textbf{if} $ GED(s_i, s_j) == g_{min} $ \textbf{then}
    \STATE $~~~~~~~~~~~~$ $ E_{pos} \leftarrow E_{pos} \cup \textbf{e}_j $ 
    \STATE $~~~~~~~~$ \textbf{else} 
    \STATE $~~~~~~~~~~~~$ $ E_{neg} \leftarrow E_{neg} \cup \textbf{e}_j $
    \STATE $~~~~~~~~$ \textbf{end}
    \STATE $~~~~$ \textbf{end for}
    \STATE $~~~~$ $\textbf{e}_c = \frac{1}{|E_{pos}|} * \sum_{\textbf{e} \in E_{pos}} \textbf{e}$  \textcolor{gray}{\# vector average}
    \STATE $~~~~$ $E_{c} \leftarrow E_{c} \cup \textbf{e}_c $
    \STATE $~~~~$ \textbf{for all} idx, $\textbf{e}_p \in E_{pos}$ \textbf{do} \textcolor{gray} {\# idx is the index of $\textbf{e}_p$ in $E_{pos}$}
    \STATE $~~~~~~~~$ $E_{pair} = \varnothing $
    \STATE $~~~~~~~~$ $sim_{p, c} = \textbf{e}_p^T\textbf{e}_c/(\tau\lVert\textbf{e}_p\rVert \lVert\textbf{e}_c\rVert)$  
    \STATE $~~~~~~~~$ $E_{pair} \leftarrow E_{pair} \cup sim_{p, c} $
    \STATE $~~~~~~~~$ \textbf{for all} $ \textbf{e}_n \in E_{neg}$ \textbf{do}
    \STATE $~~~~~~~~~~~~$ $sim_{n, c} = \textbf{e}_n^T\textbf{e}_c/(\tau\lVert\textbf{e}_n\rVert \lVert\textbf{e}_c\rVert)$
    \STATE $~~~~~~~~~~~~$ $E_{pair} \leftarrow E_{pair} \cup sim_{n, c} $
    \STATE $~~~~~~~~$ \textbf{end for}
    \STATE $~~~~~~~~$ $l_{t,idx} = -\log \frac{\exp(sim_{p, c})}{\sum_{sim_{vec} \in E_{pair}}exp(sim_{vec})}$
    \STATE $~~~~$ \textbf{end for}
    \STATE $~~~~$ $l_{t} = \sum_{idx} l_{t, idx}$
    \STATE \textbf{end for}
    \STATE $L = \frac{1}{M}\sum_{t=1}^{M} l_t + \lambda L_{reg}(E_c)$ \textcolor{gray}{\# $L_{reg}$: Eq \ref{eq_center_reg}}
    \STATE update model $f_{ccl}$ to minimize $L$
  \ENDFOR
  \STATE \textbf{return} model $f_{ccl}$
  \end{algorithmic}
\end{algorithm}

To reduce the interaction between the center vectors, we add a center vector regularization term to the loss function that forces each pair of center vectors to be orthogonal. The center vector regularization term is defined as 
\begin{equation} \label{eq_center_reg}
  L_{reg} = \frac{1}{2} \sum_{0<i,j<M}(\textbf{E}\textbf{E}^T - diag(\textbf{E}\textbf{E}^T)), \quad \textbf{E} \in \mathbb{R}^{(M*d)},
\end{equation}
where $M$ is the number of training architectures, $d$ is the dimension of the vector, $\textbf{E}$ is the matrix of center vectors with each row representing a center vector, and $i,j$ is the row and column indices of the matrix generated by the matrix multiplication $\textbf{E}\textbf{E}^T$, respectively.

In each batch of neural architectures, as one neural architecture can have the same minimum GED with several other neural architectures in this batch, we may only need to use $M$ neural architectures to optimize model $f_{ccl}$. After the pre-training, we adopt the regression loss like Eq \ref{regression_loss_precitor} to fine-tune the parameters of the neural predictor.

\subsection{Fixed Budget NPENAS} \label{nas}
NPENAS \cite{Chen2019NPENAS} combines the evolutionary search strategy with a neural predictor and utilizes the neural predictor to guide the evolutionary search strategy to explore the search space. We integrate a pre-trained neural predictor with NPENAS to illustrate the performance gains that result from applying self-supervised learning to NAS.

Since our experiments demonstrate that the neural predictor built from a self-supervised pre-trained model can significantly outperform its supervised counterpart and achieve comparable performance with a smaller training dataset, we modify the NPENAS method to utilize only a fixed search budget to carry out the neural architecture search.  

We summarize the fixed budget NPENAS in Algorithm. \ref{algo:2}, which is modified from the NPENAS \cite{Chen2019NPENAS}, and we only demonstrate the different parts.

\begin{algorithm}
  \caption{\label{algo:2} Fixed Budget NPENAS}
  \begin{algorithmic}[1]
  \STATE \textbf{Input:} initial population size $n_0$, initial population $D=\{(s_i, y_i), i=1,2,\cdots,n_0\}$, neural predictor $f$, number of the total searched architectures $total\_num$, number of the evaluated architectures (budget) to fine-tune neural predictor $ft\_num$.
  \FOR{$n$ from $n_0$ \textbf{to} $total\_num$ }
    \STATE \textbf{if} $ n \le ft\_num $ \textbf{then}
    \STATE $~~~~$ Initialize the weights of neural predictor $f$ with the weights from the pre-trained model
    \STATE  $~~~~$ Fine-tune the neural predictor $f$ with dataset $D=\{(s_i,y_i), i=1,2,\cdots,n\}$
    \STATE \textbf{end}
    \STATE Utilize the the neural predictor $f$ to guide the evolutionary neural architecture search. \textcolor{gray}{\# Detailed code can be found in the Algorithm 2 of NPENAS \cite{Chen2019NPENAS}. }
  \ENDFOR
  \STATE \textbf{Output:} $s^* = \arg \min (y_i), {(s_i,y_i)\in D}$.
  \end{algorithmic}
\end{algorithm}

\section{Experiments and Analysis} \label{experiments}
In this section, we conduct experiments to illustrate that the performance of our designed neural predictors can be significantly improved by utilizing self-supervised learning to pre-train the neural predictors' architecture embedding modules. We also demonstrate that integrating the designed pre-trained neural predictors with NPENAS is beneficial for NAS.

All the experiments are implemented in Pytorch \cite{Paszke2019PyTorchAI}. We use the implementation of GIN from the publicly available graph neural network library pytorch\_geometric \cite{FeyLenssen2019}. The code of this paper is provided at \cite{Ssnenascode}.

\subsection{Benchmark Datasets}
We perform all the experiments on the NASBench-101 \cite{Ying2019NASBench101TR} and NASBench-201 \cite{Dong2020NASBench201ET} benchmarks.

\paragraph{NASBench-101} The NASBench-101 \cite{Ying2019NASBench101TR} contains $423$k neural architectures, and each architecture is trained three times on the CIFAR-10 \cite{Krizhevsky09learningmultiple} training dataset independently. The structure of the neural architectures, as well as their validation accuracies, test accuracies corresponding to the three independently training on the CIFAR-10, are reported. The architecture in this search space is defined by DAG, utilizing nodes to represent the operations of the neural architecture and using the adjacency matrix to represent the connection of different operations. Only convolution $1 \times 1$, convolution $3 \times 3$, and max-pool $3 \times 3$ are allowed to be used to build the neural architectures. The best architecture achieves a mean test error of $5.68 \%$, and the mean test error of the architecture with the best validation error is $ 5.77 \%$. 

\paragraph{NASBench-201} The NASBench-201 \cite{Dong2020NASBench201ET} is a recently proposed NAS benchmark, and it contains $15.6$k trained architectures for image classification. Each architecture is trained once on the CIFAR-10, CIFAR-100 \cite{Krizhevsky09learningmultiple}, and ImageNet-16-120, and the ImageNet-16-120 is a down-sampled variant of ImageNet \cite{Krizhevsky2017ImageNetCW}. The structure of each architecture and its evaluation details such as training error, validation error, and test error of each architecture are reported. Each architecture is defined by a DAG, utilizing nodes to represent the feature maps and using the edges to represent the operation. The convolution $1 \times 1$, convolution $3 \times 3$, average pooling $3 \times 3$, skip connection, and zeroize operation are allowed to be used to construct the neural architectures. On the CIFAR-10, CIFAR-100, and ImageNet-16-120, the best test percentage errors are $8.48\%$, $26.49\%$, and $52.69 \%$, respectively. On the CIFAR-10, CIFAR-100, and ImageNet-16-120, architectures with the best validation error achieve the test percentage errors of $ 8.91 \%$, $26.49 \% $, and $ 53.8 \% $, respectively.

\subsection{Prediction Analysis}
\paragraph{Model Details} We first utilize the self-supervised regression learning to train the model $f_{rl}$ in Fig. \ref{fig_2} and the self-supervised central contrastive learning to train the model $f_{ccl}$ in Fig. \ref{fig_3}. The architecture embedding module consists of three sequentially connected GIN layers. The hidden layer size of the GIN layer is 32, and each GIN layer is followed with a batch normalization and a ReLU layer. The hidden dimension size of the fully connected layer of the model $f_{rl}$ and $f_{ccl}$ is 16 and 8, respectively. After self-supervised pre-training, we construct the neural predictors by connecting the pre-trained architecture embedding modules with a single fully connected layer with the hidden dimension size of 8. The neural predictors constructed by the architecture embedding modules of $f_{rl}$ and $f_{ccl}$ are denoted as SS-RL and SS-CCL, respectively.

We employ the same neural architecture encoding method as the NPENAS \cite{Chen2019NPENAS}. The architecture in the NASBench-101 is represented by a $7\times7$ upper triangle adjacency matrix and a collection of 6-dimensional one-hot encoded node features, and that in the NASBench-201 is represented by an $8 \times 8$ upper triangle matrix and several 8-dimensional one-hot encoded node features. 

\paragraph{Training Detials}
The self-supervised regression learning utilizes $90 \%$ of the neural architectures in NASBench-101 to pre-train the model $f_{rl}$, the training epoch is 300, and the batch size is 64. We employ Adam optimizer \cite{Kingma2014AdamAM} to optimize the parameters of the model $f_{rl}$, the initial learning rate is $5\mathrm{e}{-4}$, and the weight decay is $1\mathrm{e}{-4}$. A cosine learning rate schedule \cite{Loshchilov2017SGDRSG} without restart is adopted to anneal down the learning rate to zero. The training details of self-supervised regression learning on NASBench-201 are the same as for NASBench-101.

The self-supervised central contrastive learning utilizes all the architectures in NASBench-101 to pre-train the model $f_{ccl}$. The training epoch is 300, the regularization weight $\lambda$ is 0.5, and the temperature $\tau$ is $0.07$. The batch size is $140$k, the training architectures are $140$k, and we drop each epoch's last batch. When pre-train on the NASBench-201 benchmark, the batch size is $10$k, the training architectures are $1$k, and we also drop the last batch of each epoch. Other training details like the optimizer, learning rate, weight decay, and the learning rate schedule are identical with that for the self-supervised regression learning.

After pre-training, the neural architectures and their corresponding validation accuracies are used to fine-tune the neural predictors. The neural predictors are fine-tuned with an initial learning rate of $5\mathrm{e}{-5}$ and weight decay of $1\mathrm{e}{-4}$. The optimizer and the learning schedule are the same as for the self-supervised pre-training. 

\paragraph{Setup} The search budget and training epochs of neural predictors directly affect the time cost of NAS. Given that the search budget defines the size of the training dataset of the neural predictors, the search budget affects the time cost of NAS more than the number of training epochs. We conduct experiments to illustrate the performance of neural predictors under different search budgets. The supervised neural predictor, SS-RL and SS-CCL are compared under the search budgets of 20, 50, 100, 150 and 200, respectively. To illustrate the effect of training epochs, we also compare the performance of the neural predictors with different search budgets trained under 50, 100, 150, 200, 250, and 300 training epochs. The weights of the supervised neural predictor are randomly initialized. After fine-tuning, we evaluate the correlation between the validation accuracy of the neural architecture and its performance predicted by the neural predictors. The Kendall tau rank correlation is used for comparison. All the experiments results are averaged over 40 independent running using different random seeds.

\begin{figure}[ht!]
  \begin{center}
      \begin{subfigure}{0.24\textwidth}
          \includegraphics[width=0.98\linewidth, height=3.4cm]{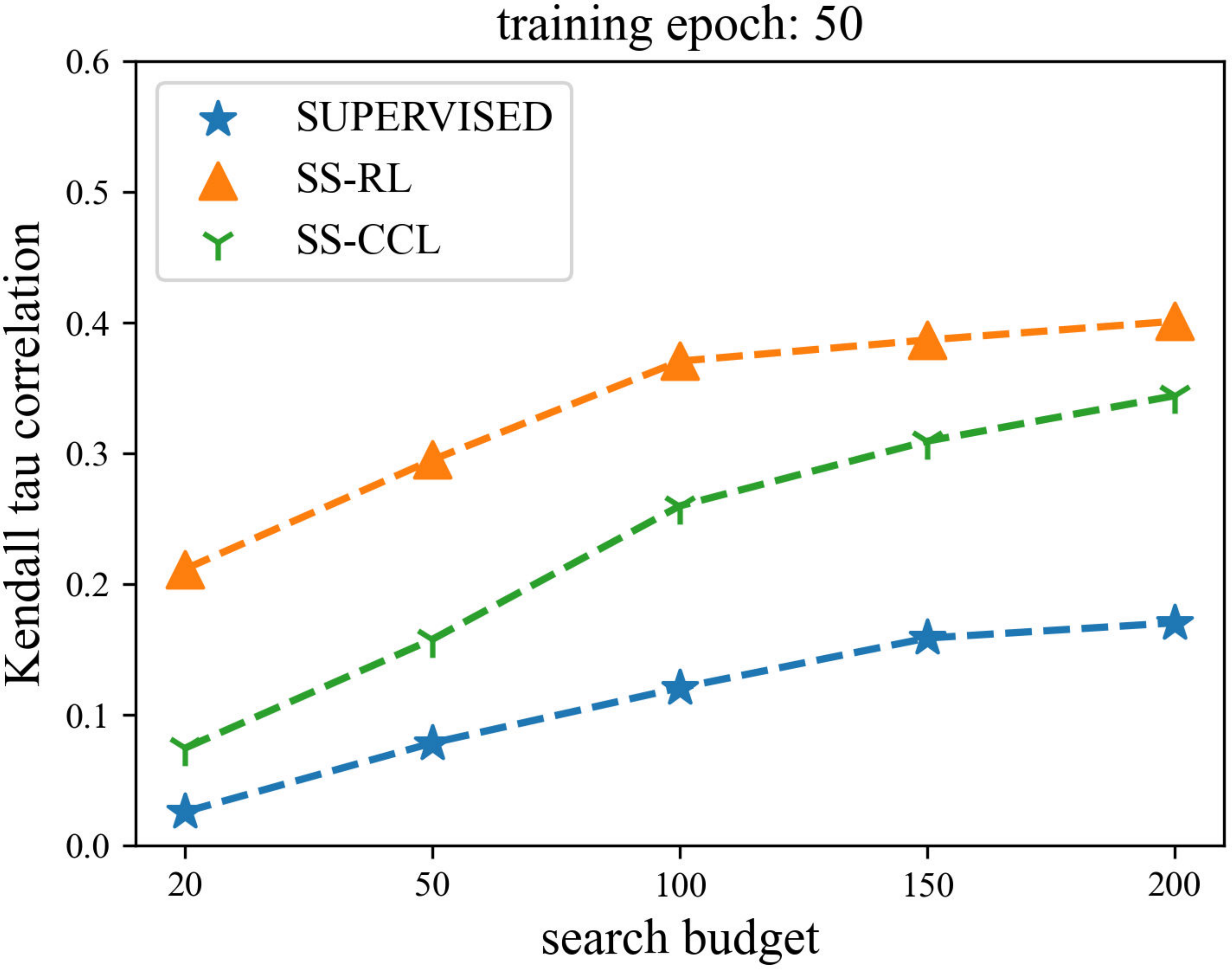}
          \caption{}
          \label{fig_4_1}
      \end{subfigure}
      \begin{subfigure}{0.24\textwidth}
          \includegraphics[width=0.98\linewidth, height=3.4cm]{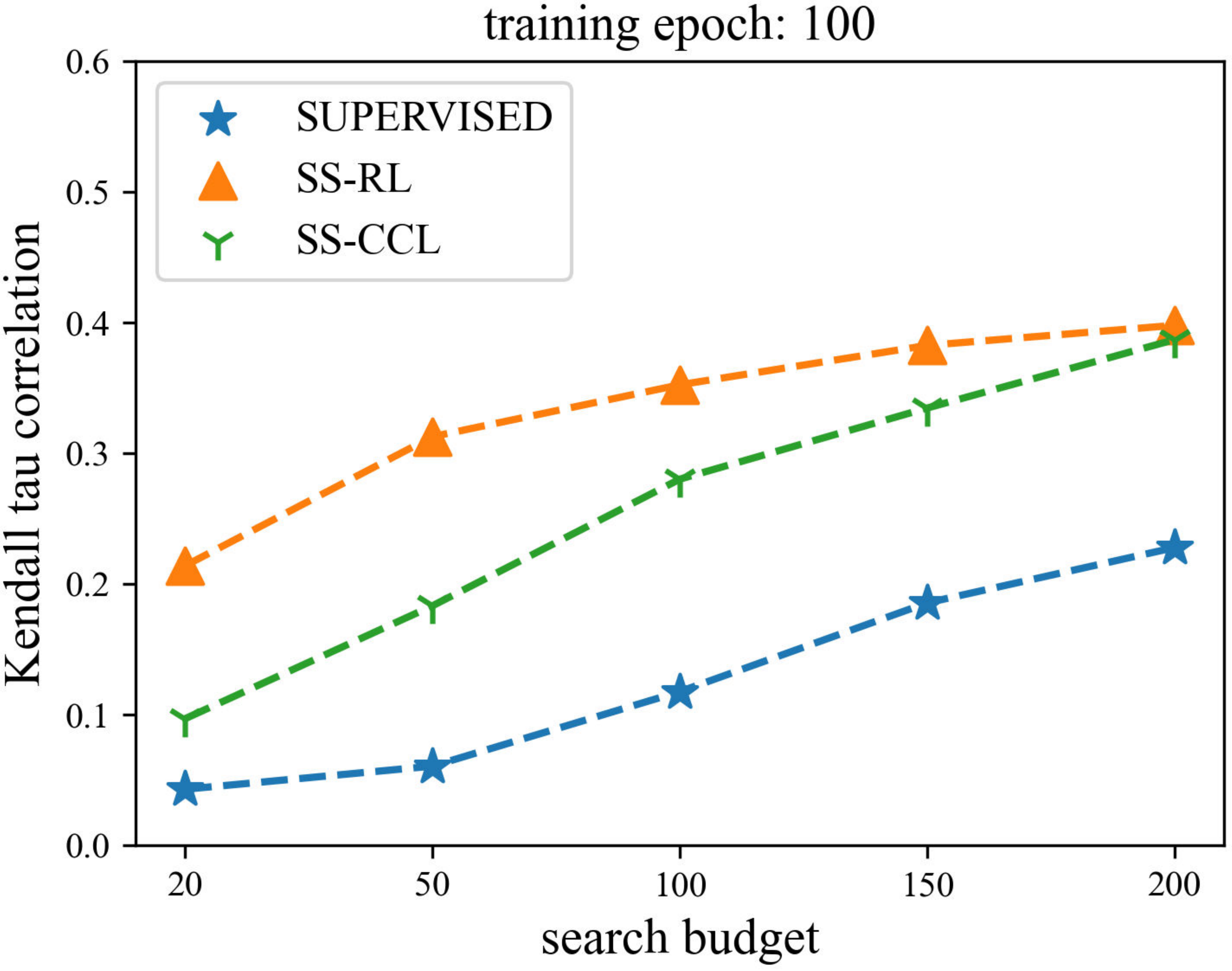}
          \caption{}
          \label{fig_4_2}
      \end{subfigure}
      \begin{subfigure}{0.24\textwidth}
        \includegraphics[width=0.98\linewidth, height=3.4cm]{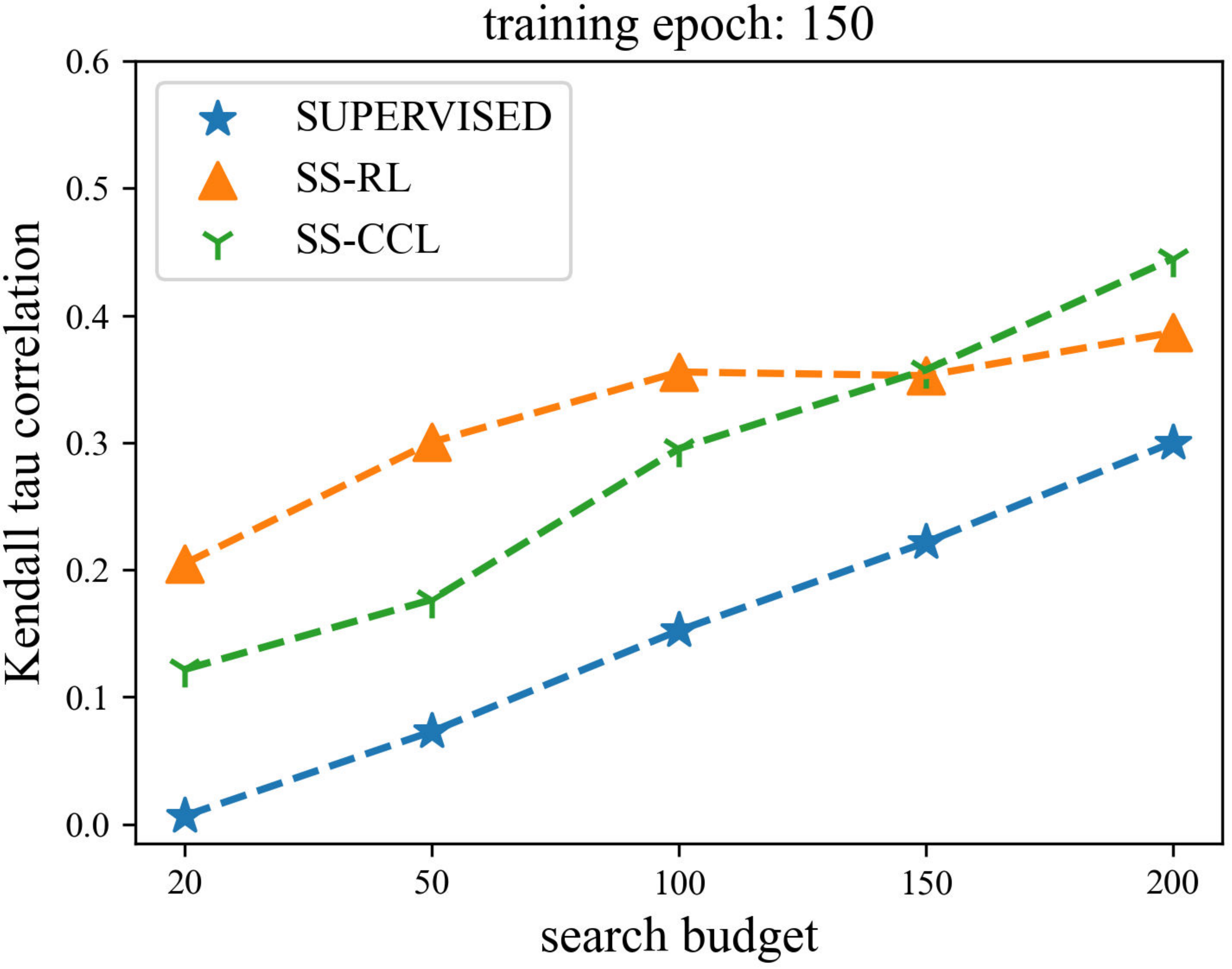}
        \caption{}
        \label{fig_4_3}
      \end{subfigure}
      \begin{subfigure}{0.24\textwidth}
        \includegraphics[width=0.98\linewidth, height=3.4cm]{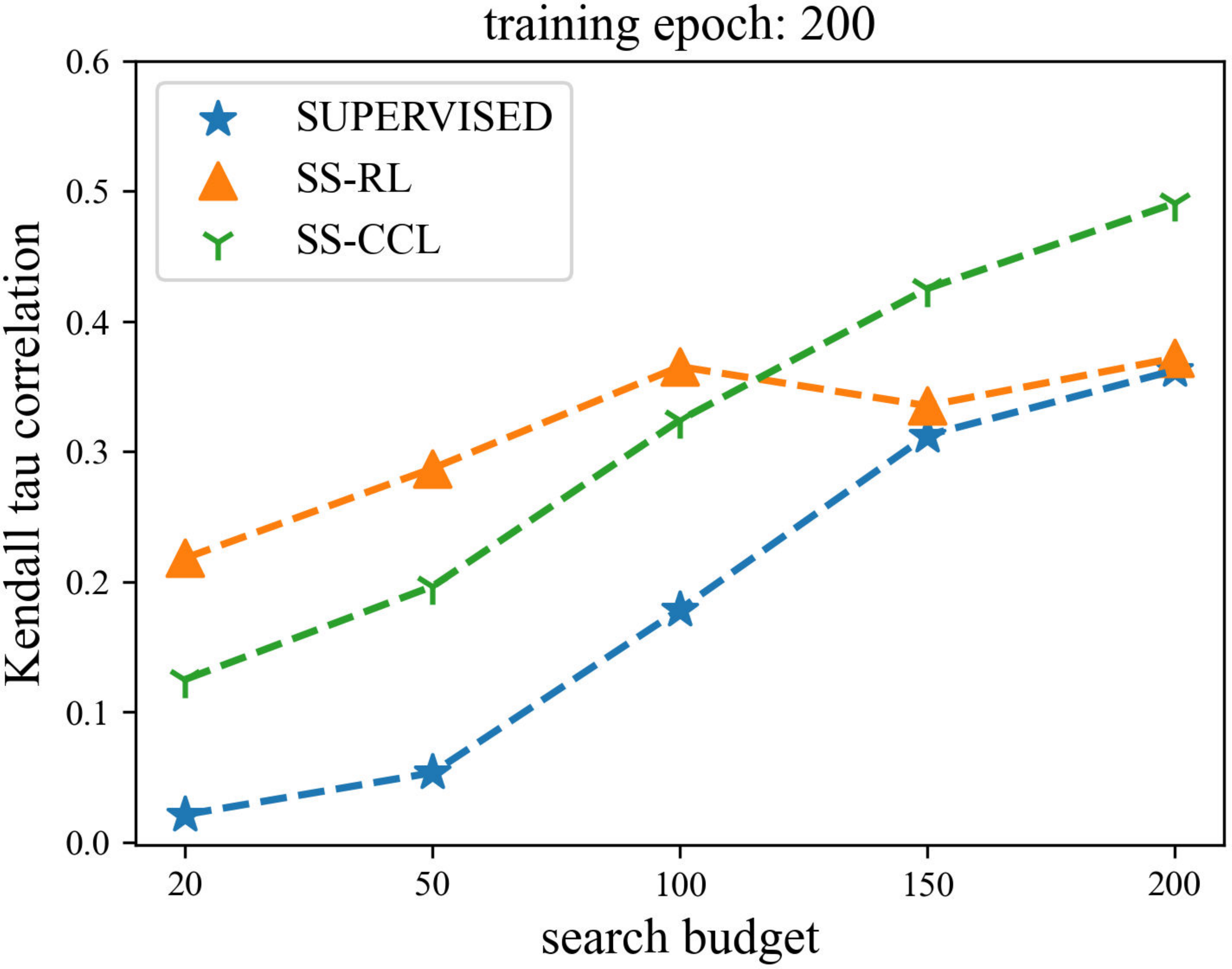}
        \caption{}
        \label{fig_4_4}
      \end{subfigure}
      \begin{subfigure}{0.24\textwidth}
        \includegraphics[width=0.98\linewidth, height=3.4cm]{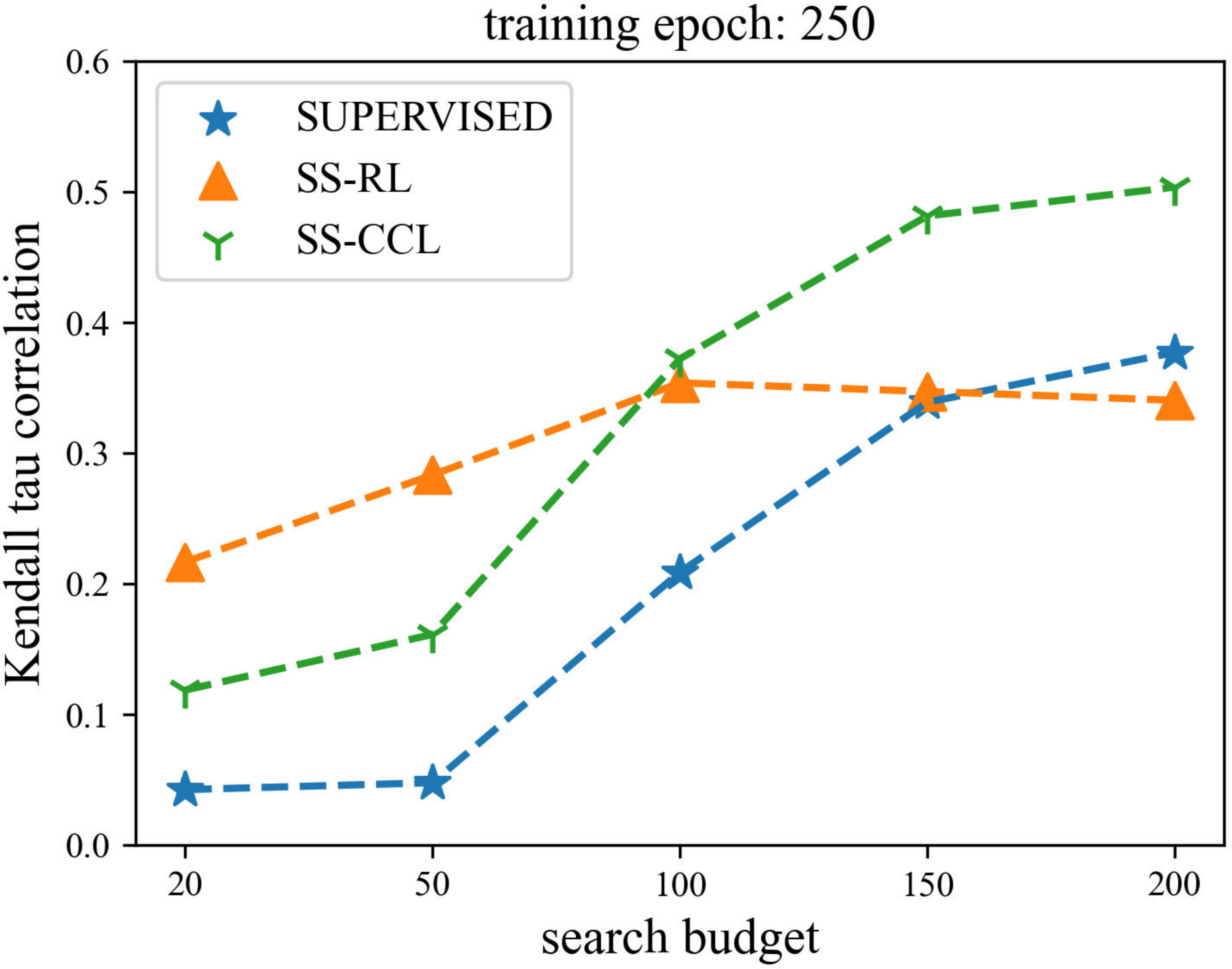}
        \caption{}
        \label{fig_4_5}
    \end{subfigure}
    \begin{subfigure}{0.24\textwidth}
      \includegraphics[width=0.98\linewidth, height=3.4cm]{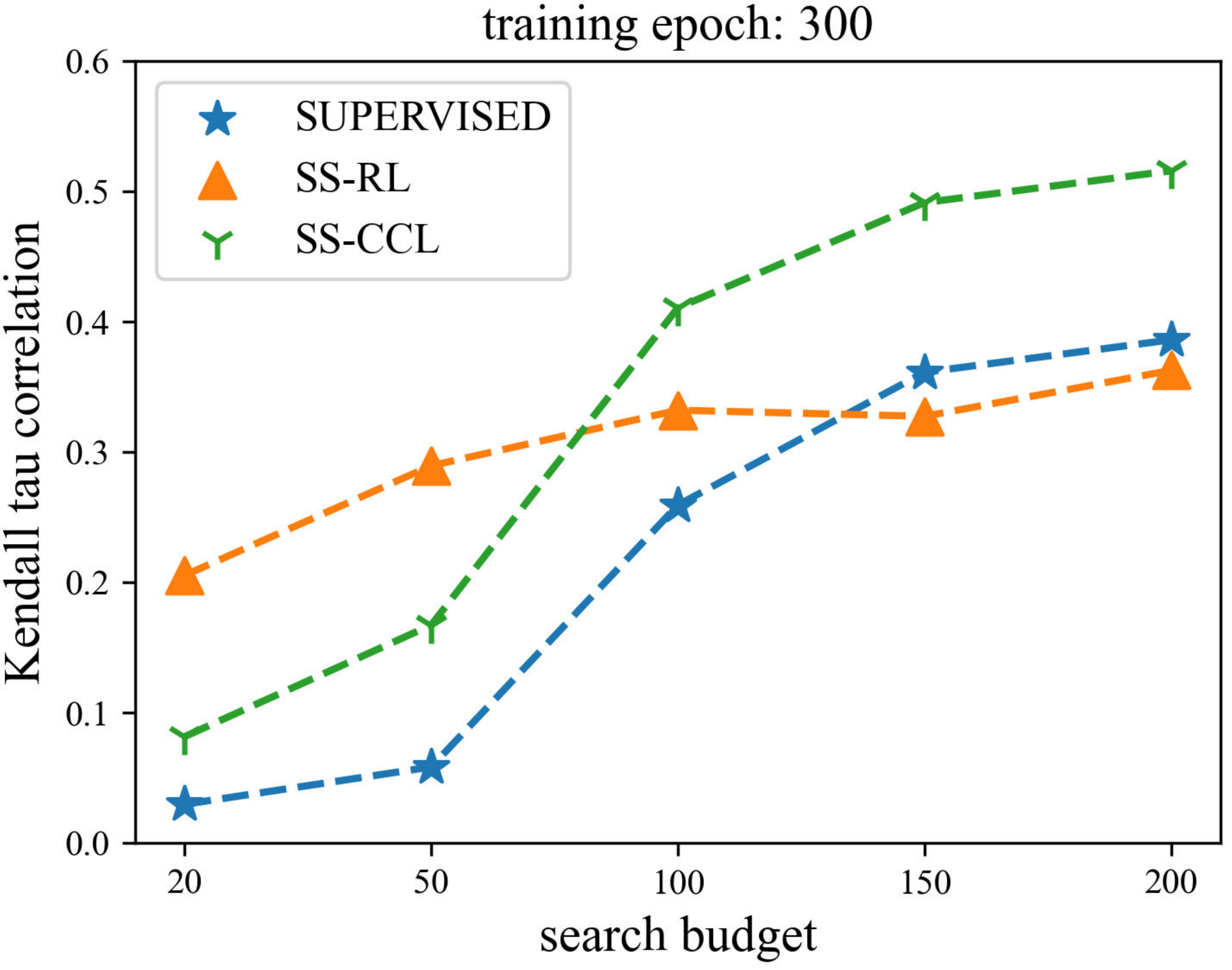}
      \caption{}
      \label{fig_4_6}
    \end{subfigure}
      \caption{Comparison of predictive performance of neural predictors on NASBench-101. (a) Training epoch 50. (b) Training epoch 100. (c) Training epoch 150. (d) Training epoch 200. (e) Training epoch 250. (f) Training epoch 300.}
      \label{fig_4}
  \end{center}
\end{figure}

\begin{figure}[ht!]
  \begin{center}
      \begin{subfigure}{0.24\textwidth}
          \includegraphics[width=0.98\linewidth, height=3.4cm]{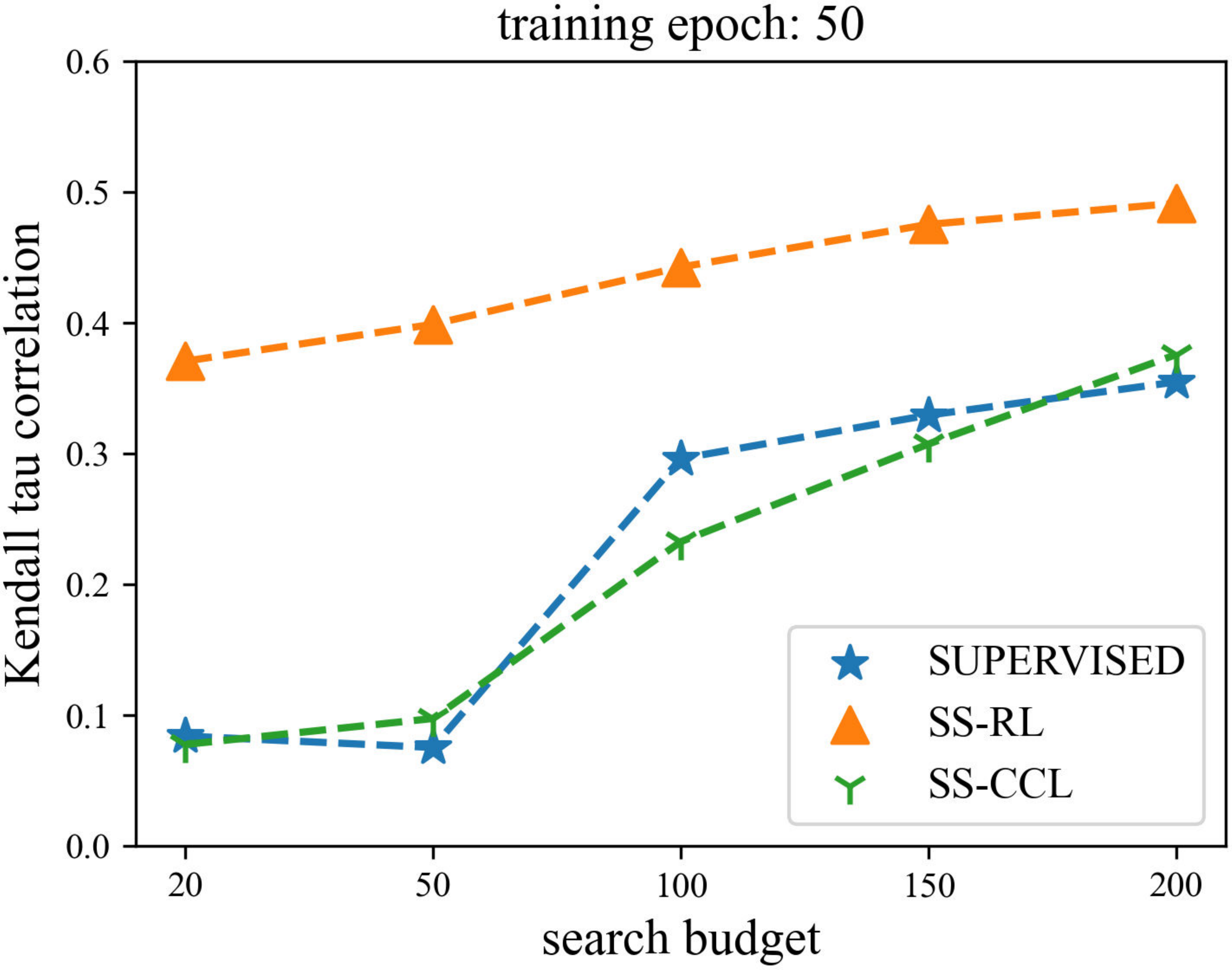}
          \caption{}
          \label{fig_5_1}
      \end{subfigure}
      \begin{subfigure}{0.24\textwidth}
          \includegraphics[width=0.98\linewidth, height=3.4cm]{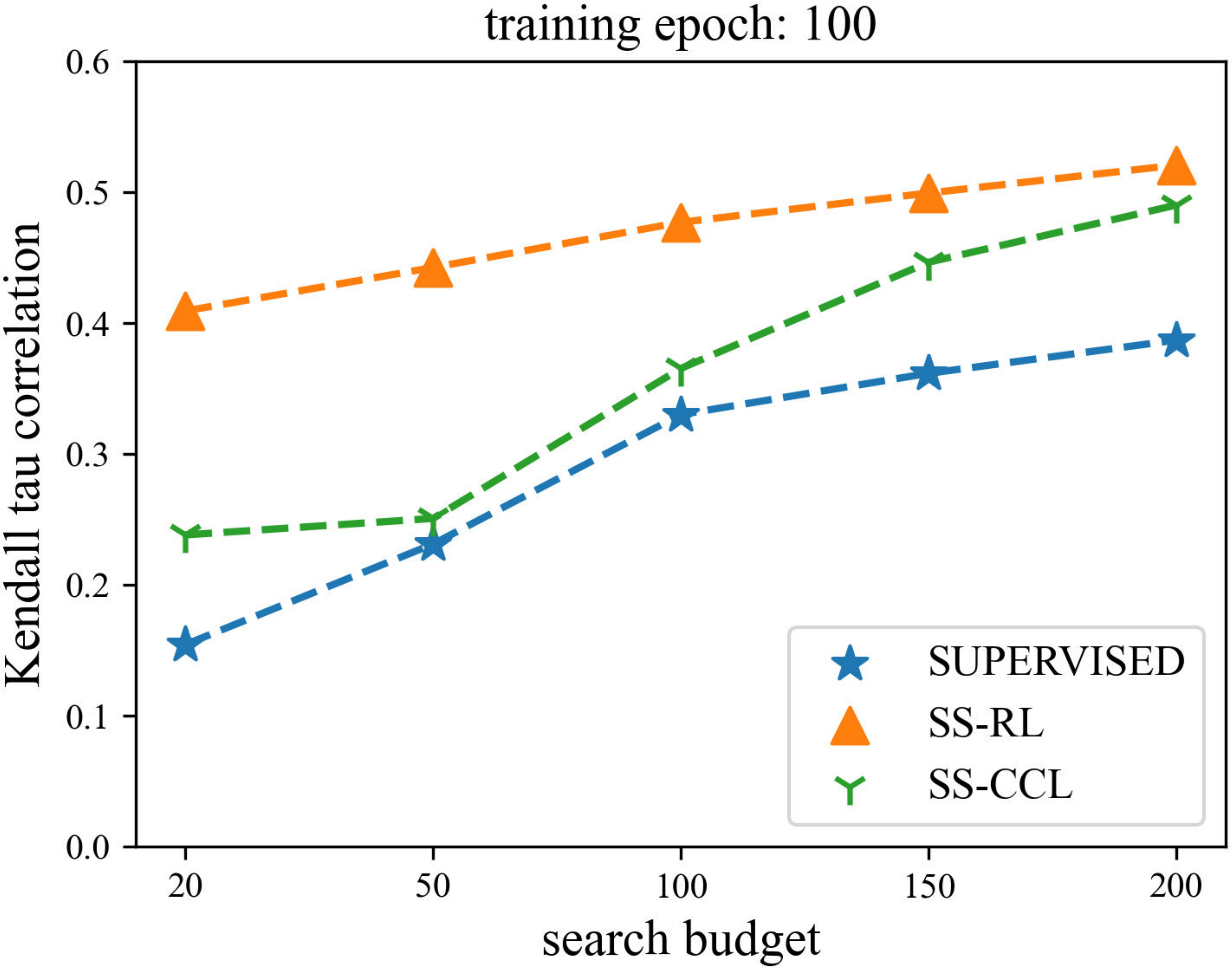}
          \caption{}
          \label{fig_5_2}
      \end{subfigure}
      \begin{subfigure}{0.24\textwidth}
        \includegraphics[width=0.98\linewidth, height=3.4cm]{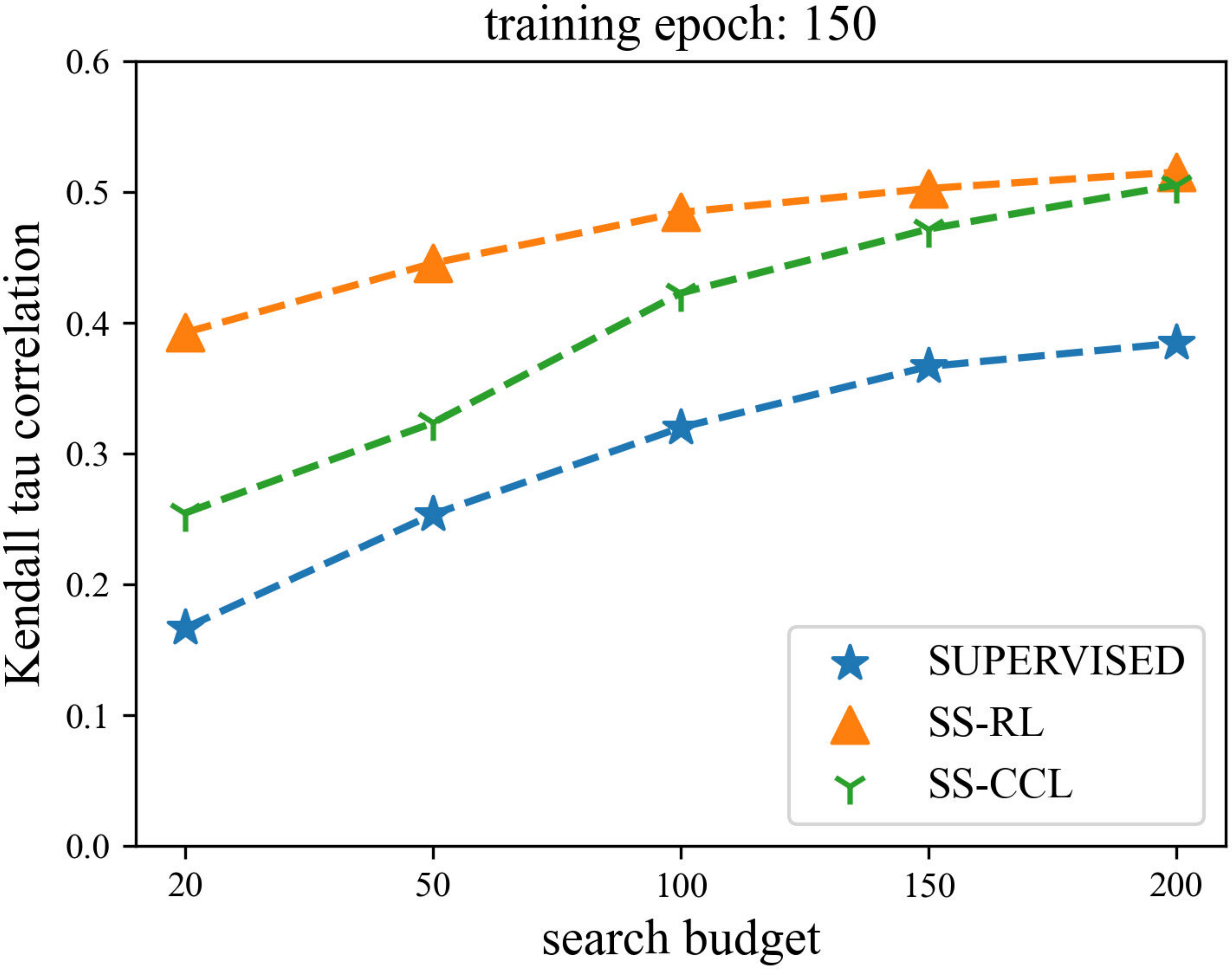}
        \caption{}
        \label{fig_5_3}
      \end{subfigure}
      \begin{subfigure}{0.24\textwidth}
        \includegraphics[width=0.98\linewidth, height=3.4cm]{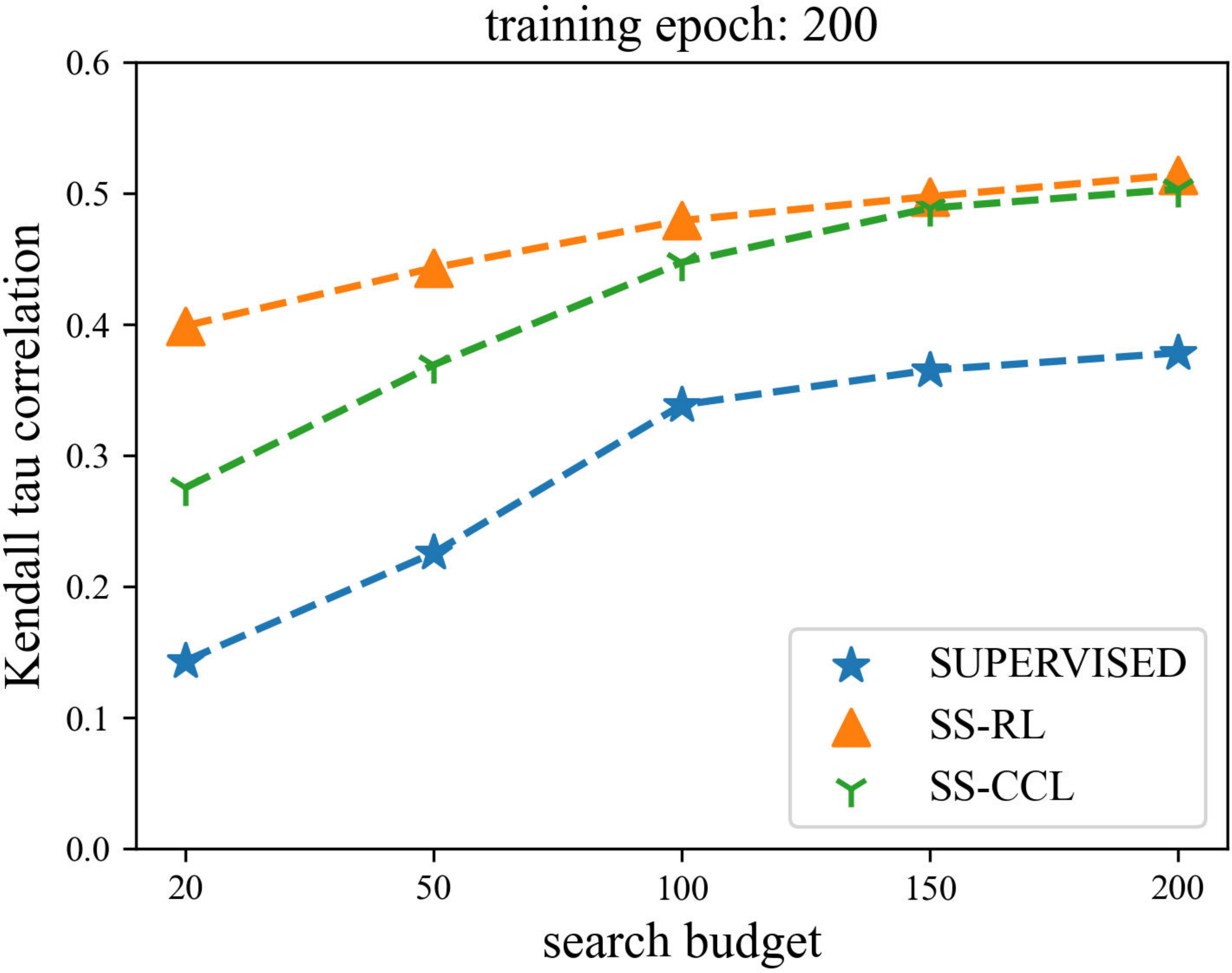}
        \caption{}
        \label{fig_5_4}
      \end{subfigure}
      \begin{subfigure}{0.24\textwidth}
        \includegraphics[width=0.98\linewidth, height=3.4cm]{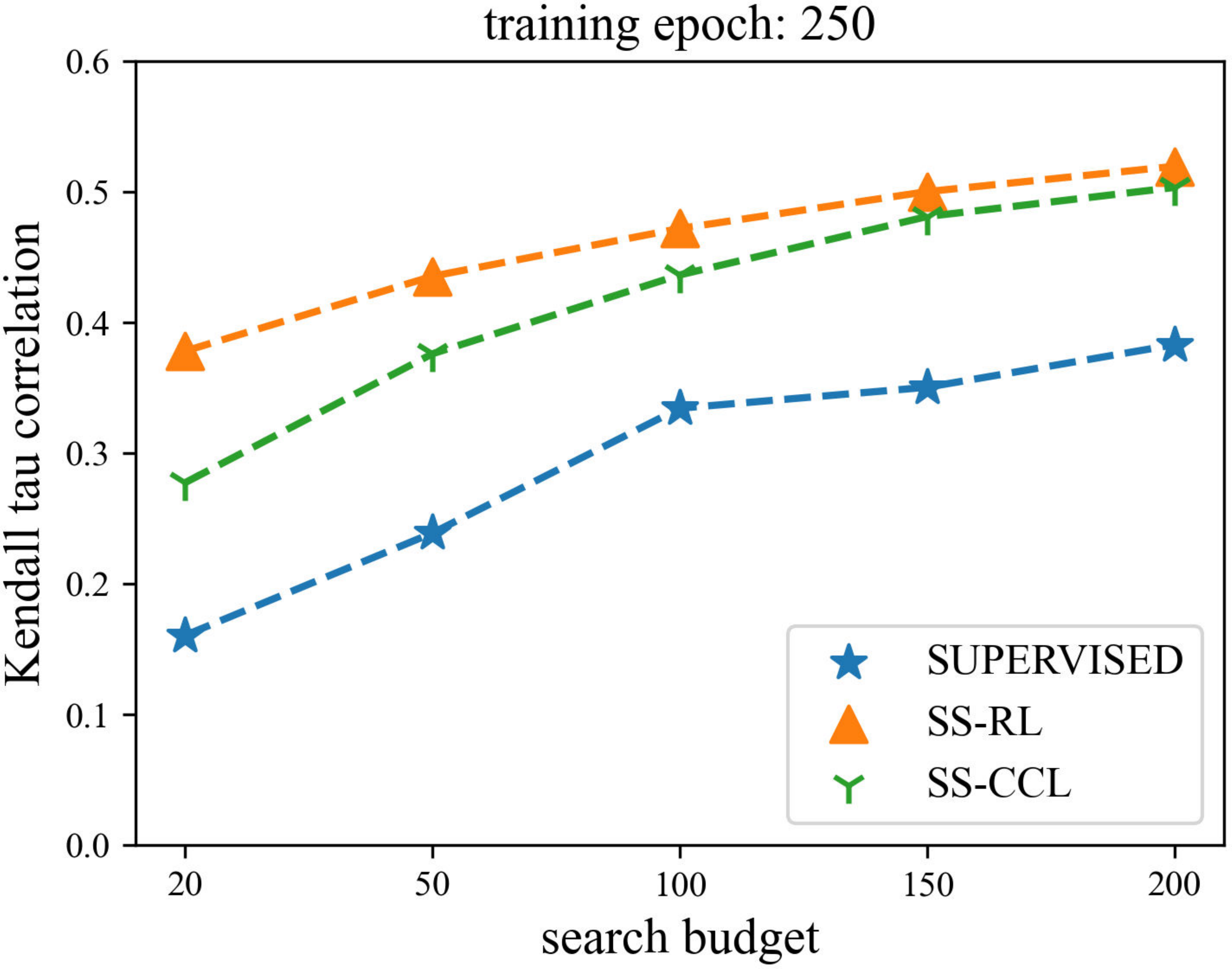}
        \caption{}
        \label{fig_5_5}
      \end{subfigure}
      \begin{subfigure}{0.24\textwidth}
        \includegraphics[width=0.98\linewidth, height=3.4cm]{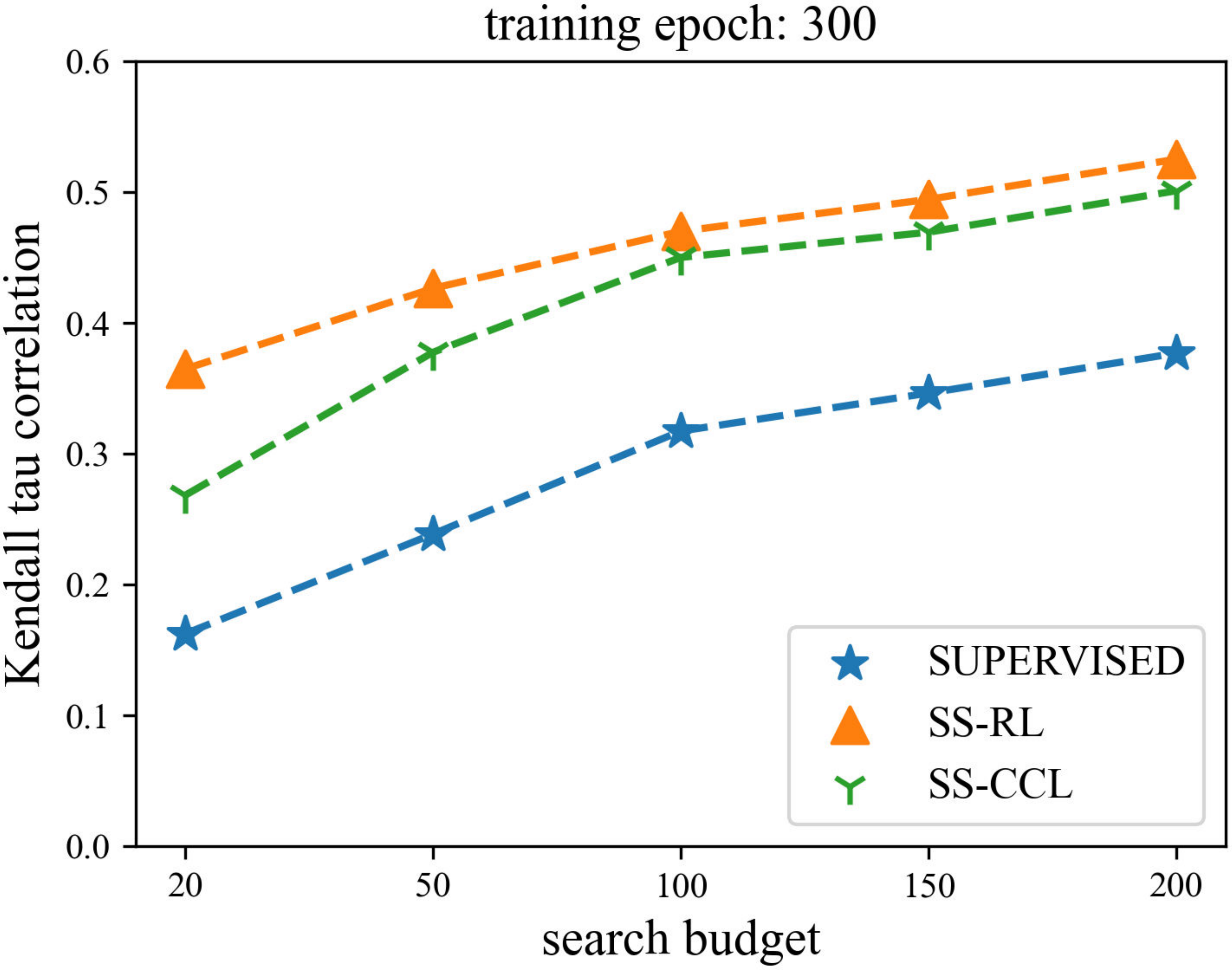}
        \caption{}
        \label{fig_5_6}
      \end{subfigure}
        \caption{Comparison of predictive performance of neural predictors on NASBench-201. (a) Training epoch 50. (b) Training epoch 100. (c) Training epoch 150. (d) Training epoch 200. (e) Training epoch 250. (f) Training epoch 300.}
        \label{fig_5}
    \end{center}
\end{figure}

\paragraph{Results} The predictive performance measurements of the neural predictors on NASBench-101 and NASBench-201 are shown in in Fig. \ref{fig_4} and Fig. \ref{fig_5}, respectively.

On the NASBench-101 search space, the predictive performance of the two pre-trained neural predictors significantly outperform the supervised neural predictor. The pre-trained neural predictor SS-RL can achieve its best performance with less search budget and training epochs, and the performance gap between SS-RL and the supervised neural predictor gradually decreases with the increasing of training epochs and search budget. When the training epochs are larger than 250, the supervised neural predictor begins to outperform SS-RL using a large search budget, e.g., more than 150. The performance of SS-CCL consistently outperforms the supervised neural predictor and increases as the training epochs increase. With the increase of training epochs, the performance of SS-CCL gradually increases. When the training epochs are larger than 150 and the search budget larger than 100, the performance of SS-CCL begins to outperform the SS-RL. When the training epochs is small, SS-RL using ten times less training neural architectures can achieve better (Fig. \ref{fig_4_1}) or comparable (Fig. \ref{fig_4_2}) performance with the supervised neural predictor. At larger training epochs, SS-CCL using twice less training neural architectures can achieve comparable (Fig. \ref{fig_4_5}) or even better performance (Fig. \ref{fig_4_6}) than the supervised neural predictor. 

On the NASBench-201 search space, the performance of SS-RL consistently outperforms SS-CCL and the supervised neural predictor. The predictive performance of SS-CCL gradually approaches that of SS-RL as training epochs increase. The performance of SS-RL using ten times less training neural architectures and only trained for 50 epochs can outperform (Fig. \ref{fig_5_1}, Fig. \ref{fig_5_6}) the supervised neural predictor trained for 300 epochs. When the training epochs are larger than 150, the performance of SS-CCL trained with four times less neural architectures can outperform (Fig. \ref{fig_5_4}, Fig. \ref{fig_5_5}, Fig. \ref{fig_5_6})the supervised neural predictor.

In summary, SS-RL can achieve its best performance with fewer training epochs, while SS-CCL requires more training epochs to achieve its best performance. As the number of training epochs increases, SS-CCL outperforms SS-RL on the NASBench-101 and tends to outperform SS-RL on the NASBench-201.

\subsection{Effect of Batch Size}
Since the number of negative pairs of the central contrastive learning is determined by the batch size, in this section, we perform experiments to investigate the effect of different batch sizes on the performance of neural predictors. We compare the prediction performance of neural predictors using the batch sizes $N$ of 10k, 40k, 70k, and 100k, and denoted the neural predictors corresponding to different $N$s as SS-CCL\_10k, SS-CCL\_40k, SS-CCL\_70k, and SS-CCL\_100k, respectively. We set the number of training architectures $M$ to be half of the batch sizes. To compare the performance of neural predictors with larger $M$, we also include SS-CCL that is pre-trained with the batch size of 140k and the number of training architectures of 140k, and denoted it as SS-CCL\_140k. All the results are averaged over 40 independently running, and each running uses a different seed.

\begin{figure}[ht!]
  \begin{center}
      \begin{subfigure}{0.24\textwidth}
          \includegraphics[width=0.98\linewidth, height=3.5cm]{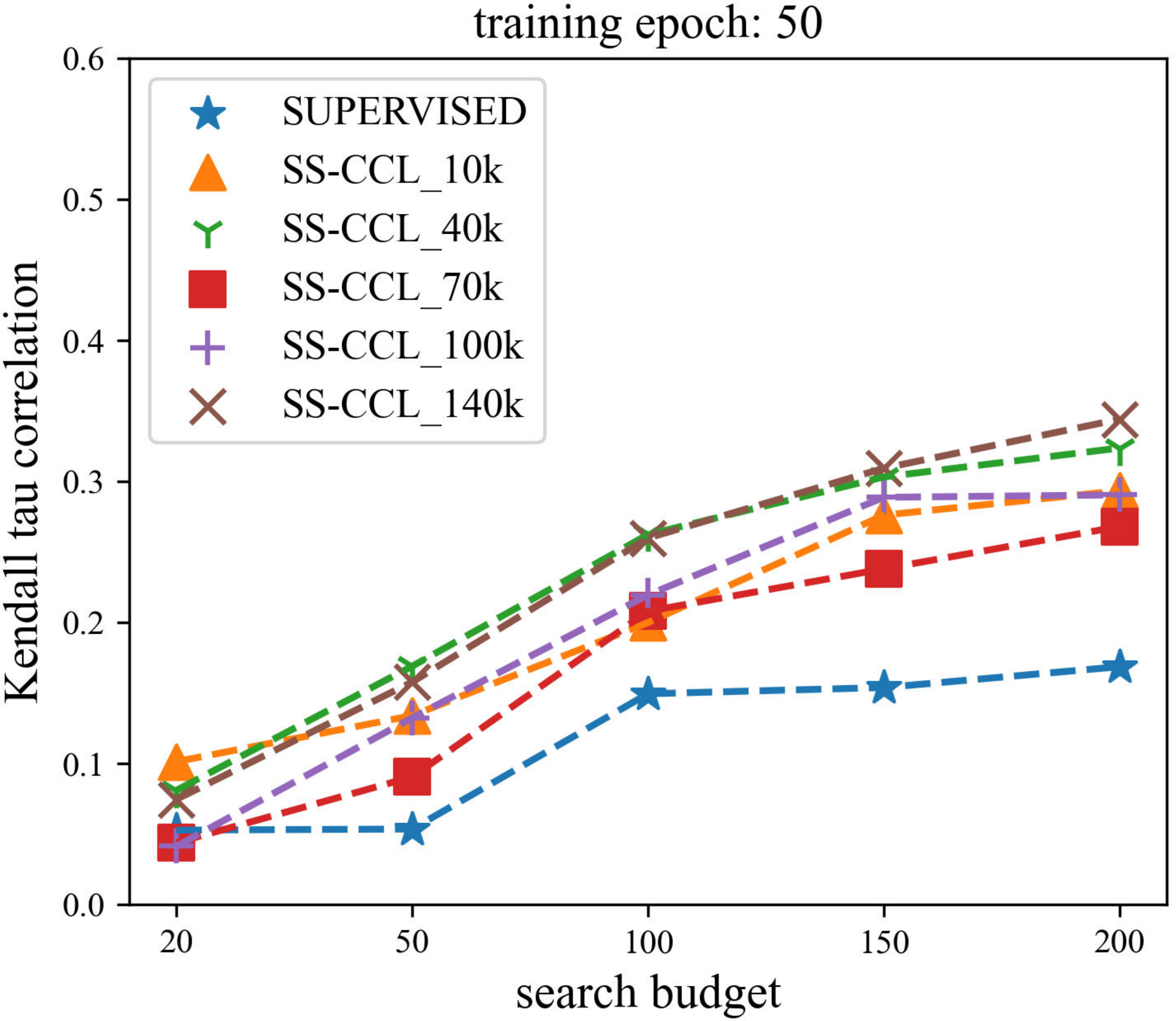}
          \caption{}
          \label{fig_6_1}
      \end{subfigure}
      \begin{subfigure}{0.24\textwidth}
          \includegraphics[width=0.98\linewidth, height=3.5cm]{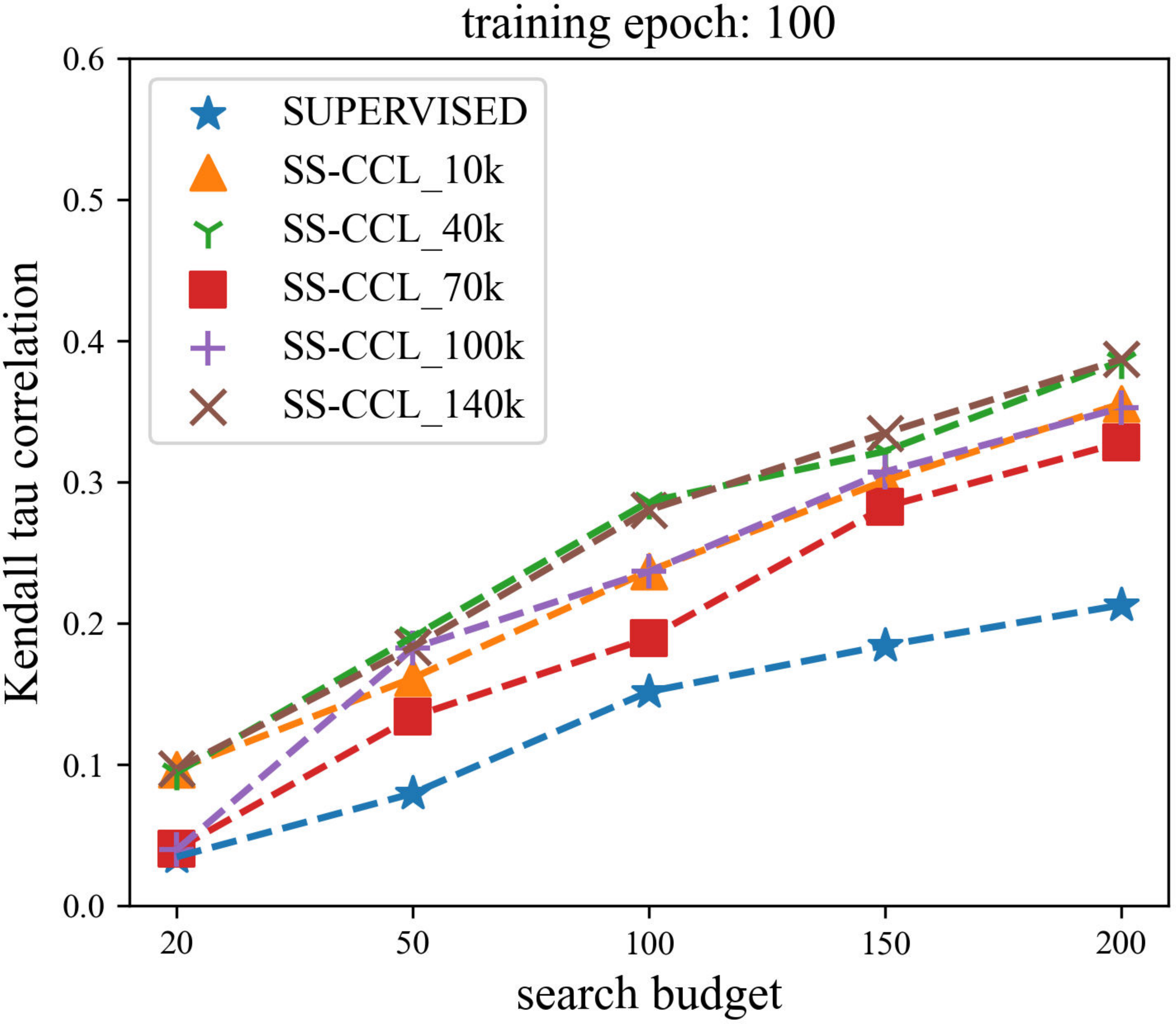}
          \caption{}
          \label{fig_6_2}
      \end{subfigure}
      \begin{subfigure}{0.24\textwidth}
        \includegraphics[width=0.98\linewidth, height=3.5cm]{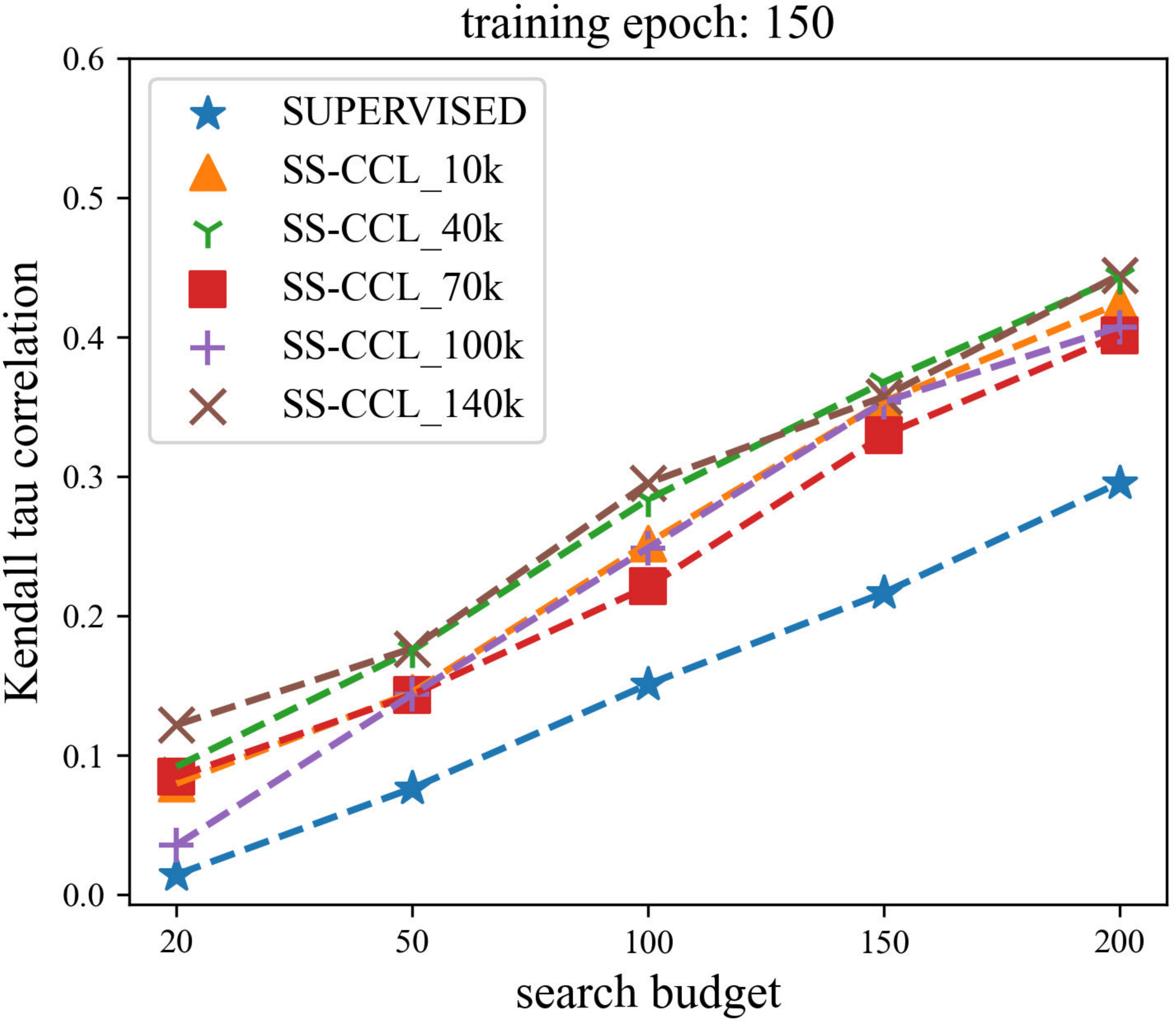}
        \caption{}
        \label{fig_6_3}
      \end{subfigure}
      \begin{subfigure}{0.24\textwidth}
        \includegraphics[width=0.98\linewidth, height=3.7cm]{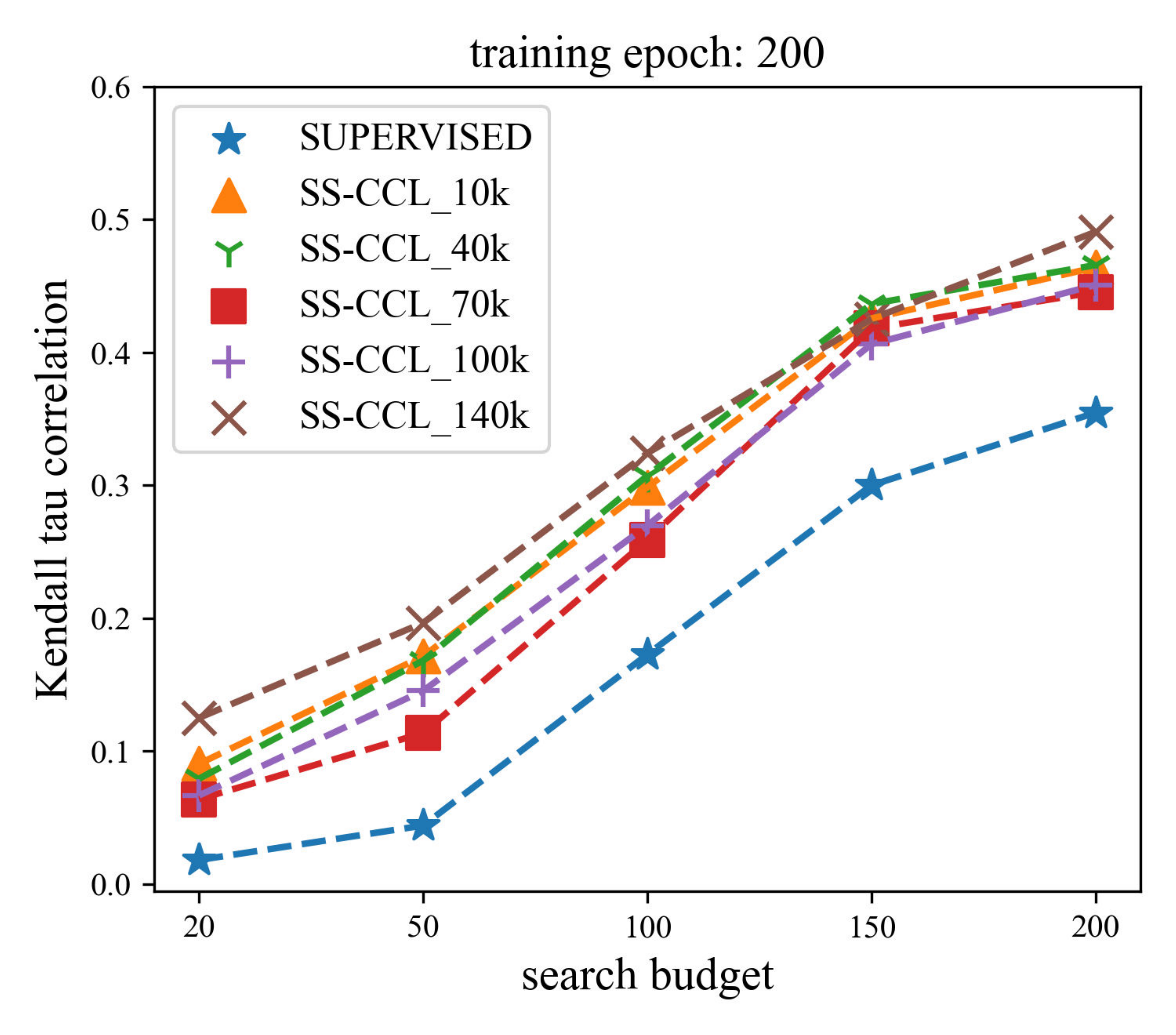}
        \caption{}
        \label{fig_6_4}
      \end{subfigure}
      \begin{subfigure}{0.24\textwidth}
        \includegraphics[width=0.98\linewidth, height=3.5cm]{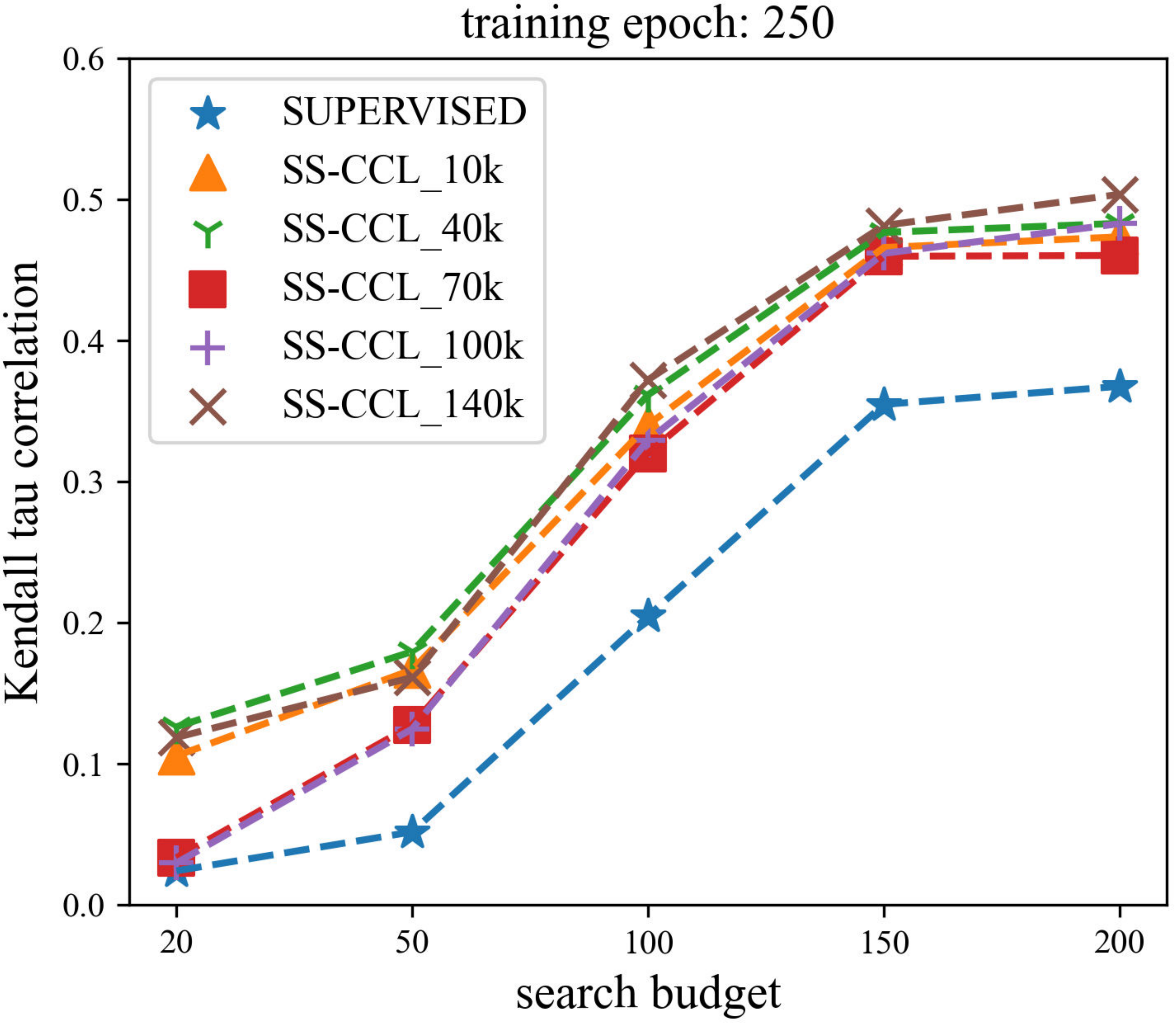}
        \caption{}
        \label{fig_6_5}
      \end{subfigure}
      \begin{subfigure}{0.24\textwidth}
        \includegraphics[width=0.98\linewidth, height=3.5cm]{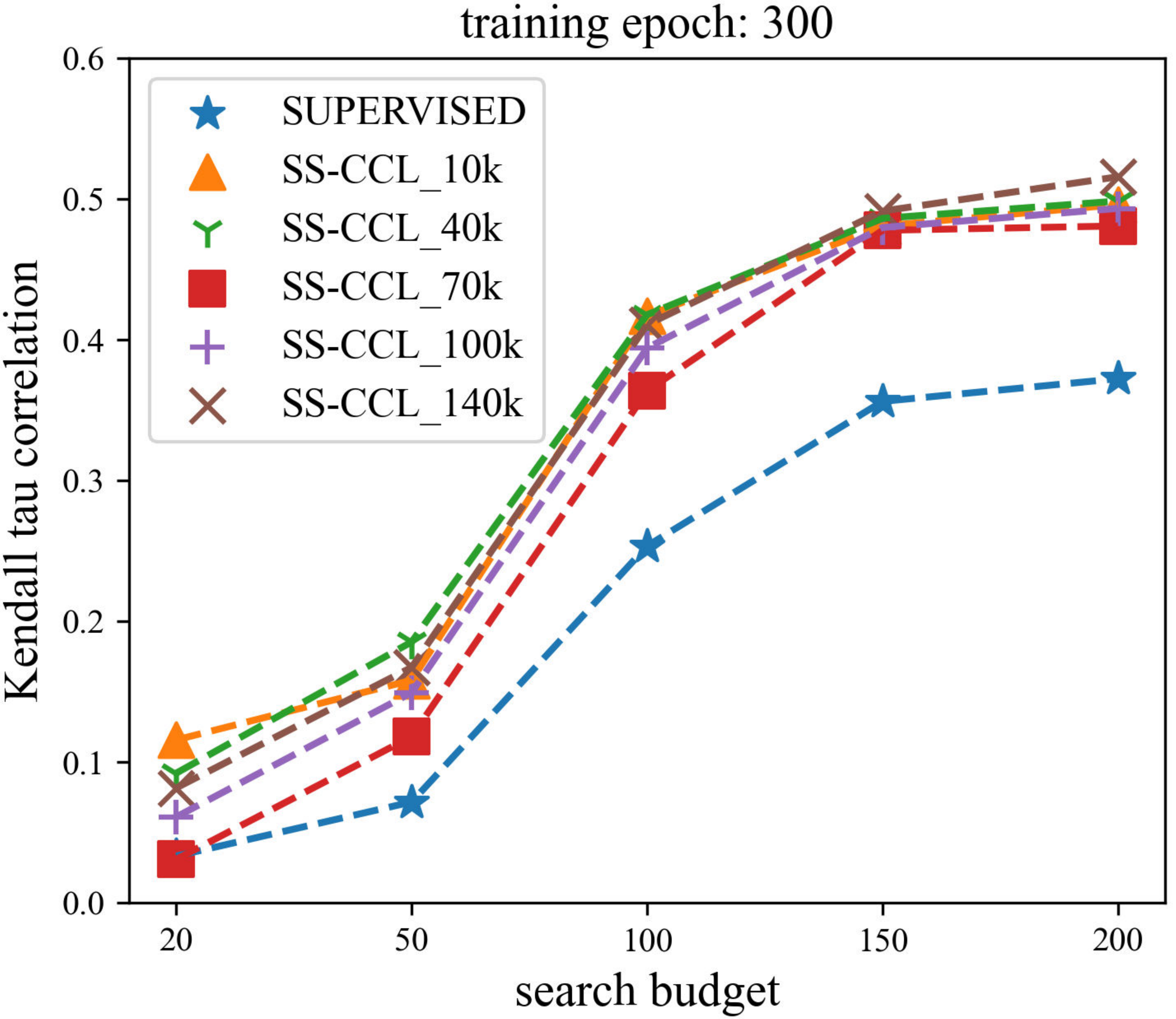}
        \caption{}
        \label{fig_6_6}
      \end{subfigure}
      \caption{Comparison of the predictive performance of the neural predictor pre-trained by different batch size $N$ on NASBench-101. (a) Training epoch 50. (b) Training epoch 100. (c) Training epoch 150. (d) Training epoch 200. (e) Training epoch 250. (f) Training epoch 300.}
      \label{fig_6}
  \end{center}
\end{figure}

As shown in Fig. \ref{fig_6}, when we only take the neural predictors with the same training architectures $M$ (excluding SS-CCL\_140k) into consideration, SS-CCL\_40k is consistently better than other neural predictors, and SS-CCL\_70k tends to be the worst compared with other pre-trained neural predictors. The above result is different from the findings in the contrastive learning for image classification \cite{DBLP:journals/corr/abs-2002-05709} that the performance is consistently increasing with large batch size and more training epochs. The predictive performance of SS-CCL-140k has a slightly better performance than SSL-CCL\_40k.
Because the large batches generate more negative pairs, making the contrastive learning more difficult, we conjecture that as the batch size $N$ in Algorithm 1 increases, the number of training architecture $M$ should also increase. The predictive performance of all the pre-trained neural predictors continuously increases with increasing training epochs, and the performance gap between the different pre-trained neural predictors decreases with the increasing of training epochs and search budget.

\subsection{Fixed Budget NPENAS}
\paragraph{Setup} We integrate our pre-trained neural predictors with NPENAS \cite{Chen2019NPENAS} and denote the integration of neural predictors SS-RL and SS-CCL with NPENAS as NPENAS-SSRL and NPENAS-SSCCL, respectively. The fixed budget version of NPENAS-SSRL and NPENAS-SSCCL are denoted as NPENAS-SSRL-FIXED and NPENAS-SSCCL-FIXED, respectively. We adopt the same experimental setting as NPENAS, and compare with the random search (RS) \cite{Li2019RandomSA}, regularized evolutionary (REA) \cite{Real2018RegularizedEF}, BANANAS \cite{White2019BANANASBO} with path-based encoding (BANANAS-PE), BANANAS with adjacency matrix encoding (BANANAS-AE), BANANAS with position-aware path-based encoding (BANANAS-PAPE), NPENAS-NP \cite{Chen2019NPENAS}, and NPENAS-NP with fixed search budget (NPENAS-NP-FIXED). Each algorithm is given a search budget of 150 and 100 on the NASBench-101 and NASBench-201 search space, respectively. All the experiment results are averaged over 600 independent trails, every update of the population, each algorithm returns the architecture with the lowest validation error so far and reports its test error, so there are 15 or 10 best architectures in total. We also compare with the arc2vec \cite{DBLP:journals/corr/abs-2006-06936} that is a recently proposed unsupervised representation learning for NAS, and directly adopt its reported results. As the search strategies employ the neural architectures' validation error to explore the search space, a reasonable best performance of NAS is the test error of the neural architecture that has the best validation error in the search space, which is denoted as the \textit{ORACLE} baseline \cite{DBLP:conf/eccv/WenLCLBK20}. We use the \textit{ORACLE} baseline as the upper bound of performance.

\subsubsection{NAS Results on NASBench-101}
\begin{figure}[ht!]
  \begin{center}
      \includegraphics[width=0.98\linewidth, height=6.5cm]{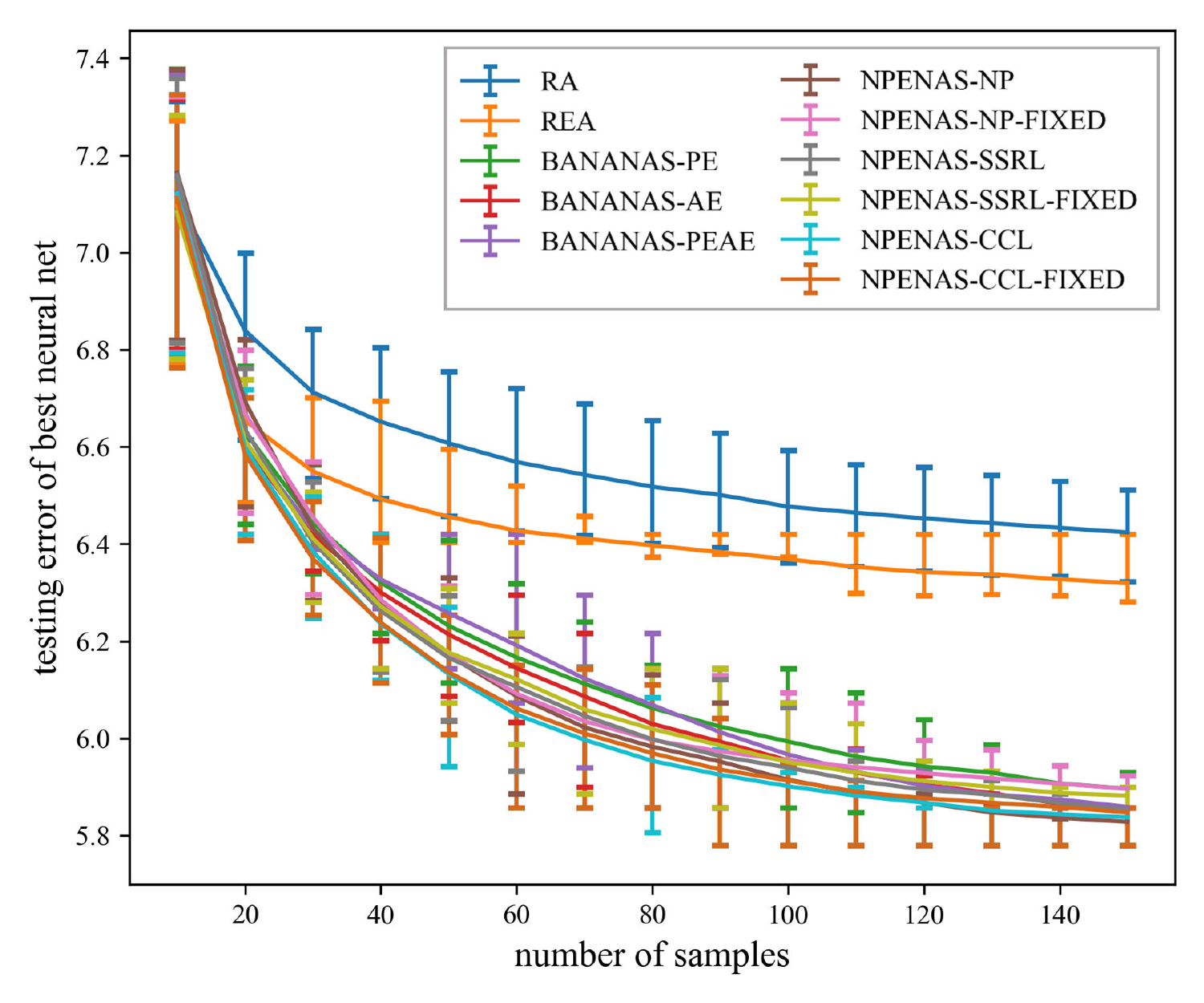}
      \caption{Performance of NAS algorithms on the NASBench-101 benchmark.}
      \label{fig_7}
  \end{center}
\end{figure}

The comparison of different algorithms is illustrated in Fig. \ref{fig_7}, and we also demonstrate the quantitative comparison of algorithms in Table \ref{table_1}. As illustrated in Fig. \ref{fig_7}, the performance of NPENAS-SSCCL is slightly better than NPENAS-SSRL, and NPENAS-NP achieves the best performance. The proposed position-aware path-based encoding is an efficient and effective encoding scheme. The performance of BANANAS \cite{White2019BANANASBO} with position-aware path-based encoding is better than BANANAS with the path-based encoding. We utilize the position-aware path-based encoding to filter out the isomorphic graphs, while NPENAS employs the path-based encoding to filter out isomorphic graphs. Due to this difference, the performance of NPENAS-NP shown in Table \ref{table_1} improves from $5.86 \%$ to $5.83 \%$. 
Table \ref{table_1} also shows our proposed self-supervised pre-trained neural predictors using a small search budget performs better than the unsupervised \textit{arch2vec}. 

\begin{table*}[ht!]
    \small
    \begin{center}
    \caption{Performance comparison of NAS algorithms on NASBench-101}
    \label{table_1}
    \begin{threeparttable}[b]
    \begin{tabular}{cccccc}
        \hline
        Methods  & \thead{Search \\ Budget}  & Test Err (\%) Avg & \thead{Architecture \\ Embedding} & Search Method \\ \hline
        RA \cite{Li2019RandomSA} &  150  &  6.42 $\pm$ 0.2 & -- & Random Search \\
        REA \cite{Real2018RegularizedEF} &  150  &  6.32 $\pm$ 0.2 & Discrete & Evolution \\ 
        BANANAS-PE \cite{White2019BANANASBO} & 150 & 5.9 $\pm$ 0.15 & Supervised & Bayesian Optimization \\ 
        BANANAS-AE \cite{White2019BANANASBO}  & 150 & 5.85 $\pm$ 0.14 & Supervised & Bayesian Optimization \\
        BANANAS-PEAE \cite{White2019BANANASBO} & 150 & 5.86 $\pm$ 0.14 & Supervised & Bayesian Optimization \\
        NPENAS-NP \cite{Chen2019NPENAS} & 150 & \textbf{5.83 $\pm$ 0.11} & Supervised & Evolution \\
        NPENAS-NP-FIXED  \cite{Chen2019NPENAS} & 90$^\dagger$ & 5.9 $\pm$ 0.16 & Supervised & Evolution \\ 
        \textit{arch2vec}-RL \cite{DBLP:journals/corr/abs-2006-06936} & 400 & 5.9 & Unsupervised & REINFORCE \\ 
        \textit{arch2vec}-BO \cite{DBLP:journals/corr/abs-2006-06936} & 400 & 5.95 & Unsupervised & Bayesian Optimization \\ \hline
        NPENAS-SSRL  & 150 & 5.85 $\pm$ 0.13 & Self-supervised & Evolution \\
        NPENAS-SSRL-FIXED  & 90$^\dagger$ & 5.88 $\pm$ 0.16 & Self-supervised & Evolution \\
        NPENAS-SSCCL  & 150 & \textbf{5.84 $\pm$ 0.12}  & Self-supervised & Evolution \\
        NPENAS-SSCCL-FIXED  & 90$^\dagger$ & 5.85 $\pm$ 0.13 & Self-supervised & Evolution \\ \hline
    \end{tabular}
    \begin{tablenotes}
      \item[$\dagger$] {\footnotesize The neural predictor is trained with 90 neural architectures, while the NPENAS algorithm needs 150 neural architectures.}
  \end{tablenotes}
    \end{threeparttable}
\end{center}
\end{table*}

As shown in Table \ref{table_1}, the performance of NPENAS-NP has a large drop after switching to the fixed version, while NPENAS-SSRL and NPENAS-SSCCL only have a slight drop. We compare the performance of NPENAS-SSRL and NPENAS-SSCCL under differ search budget, and the results are shown in Table \ref{table_2}. The performance of NPENAS-SSRL continuously increases with the increase of the search budget, while the NPENAS-SSCCL only using 80 neural architectures to achieve its best performance. From the above findings, the neural predictor SS-CCL is better than SS-RL when applying to the NPENAS.

\begin{table}[ht!]
  \small
  \begin{center}
  \caption{Comparison of NAS algorithms on NASBench-101}
  \label{table_2}
  \begin{threeparttable}[b]
  \begin{tabular}{cccccc}
      \hline
      Methods  & \thead{Search \\ Budget$^\dagger$}  & Test Err (\%) Avg \\ \hline
      NPENAS-SSRL &  20  &  6.1 $\pm$ 0.27 \\
      NPENAS-SSRL &  50  &  5.93 $\pm$ 0.19 \\ 
      NPENAS-SSRL & 80 & 5.88 $\pm$ 0.16  \\ 
      NPENAS-SSRL  & 110 & 5.86 $\pm$ 0.15  \\  
      NPENAS-SSRL  & 150 & 5.85 $\pm$ 0.13 \\  \hline
      NPENAS-SSCCL & 20 & 6.0 $\pm$ 0.2  \\
      NPENAS-SSCCL & 50 & 5.87 $\pm$ 0.15  \\
      NPENAS-SSCCL & 80 & \textbf{5.84 $\pm$ 0.13}  \\ 
      NPENAS-SSCCL  & 110 & \textbf{5.84 $\pm$ 0.12}  \\ 
      NPENAS-SSCCL  & 150 & \textbf{5.84 $\pm$ 0.12} \\  \hline
  \end{tabular}
  \begin{tablenotes}
    \item[$\dagger$] {\footnotesize The neural predictor is trained with the given number of evaluated neural architectures, while the NPENAS algorithm needs 150 evaluate architectures.}
\end{tablenotes}
  \end{threeparttable}
\end{center}
\end{table}

\subsubsection{NAS Results on NASBench-201}
\begin{figure}[ht!]
  \begin{center}
      \includegraphics[width=0.98\linewidth, height=6.5cm]{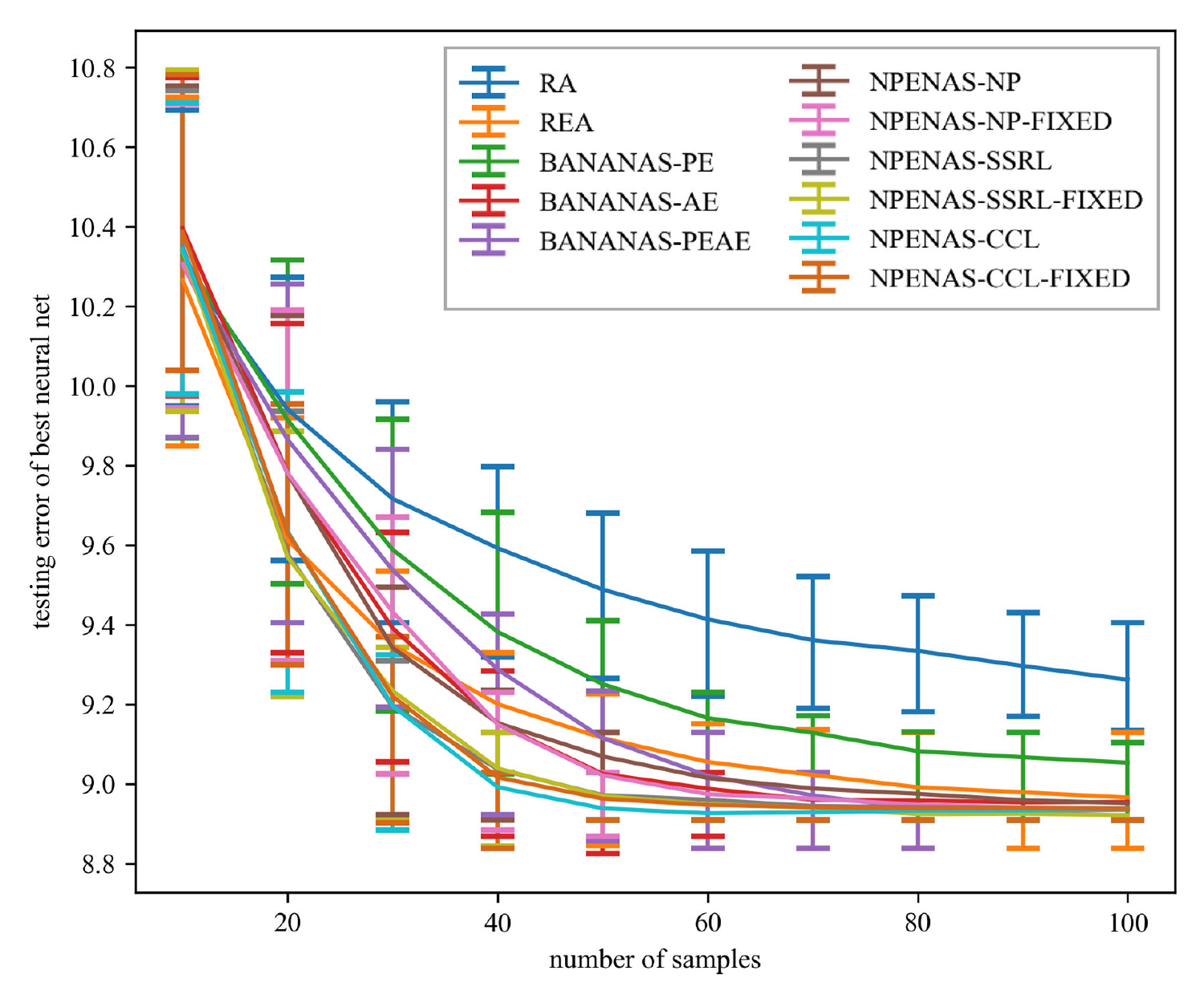}
      \caption{Performance of NAS algorithms on CIFAR-10 on NASBench-201.}
      \label{fig_8}
  \end{center}
\end{figure}

\begin{figure}[ht!]
  \begin{center}
      \includegraphics[width=0.98\linewidth, height=6.5cm]{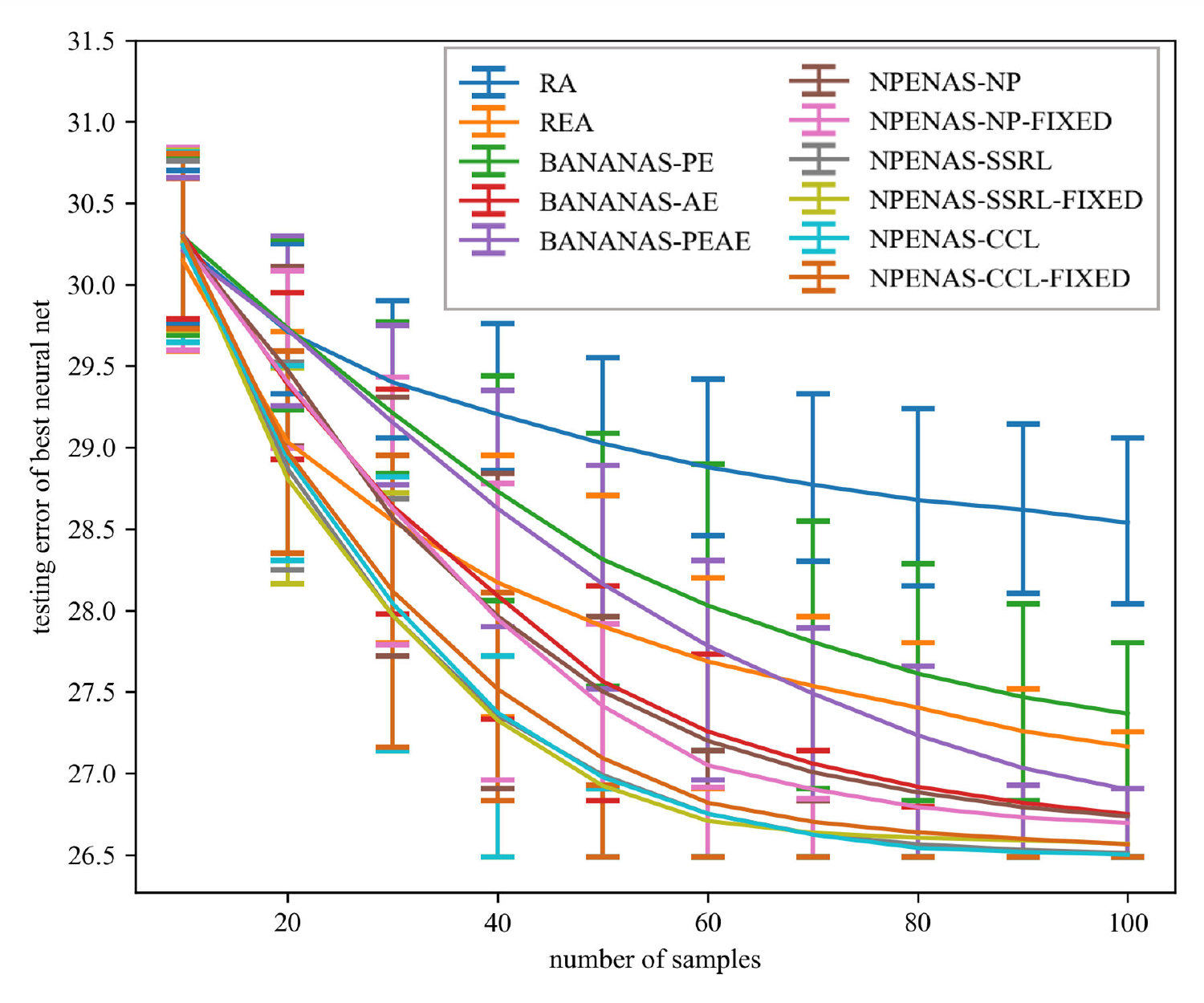}
      \caption{Performance of NAS algorithms on CIFAR-100 on NASBench-201.}
      \label{fig_9}
  \end{center}
\end{figure}

\begin{figure}[ht!]
  \begin{center}
      \includegraphics[width=0.98\linewidth, height=6.5cm]{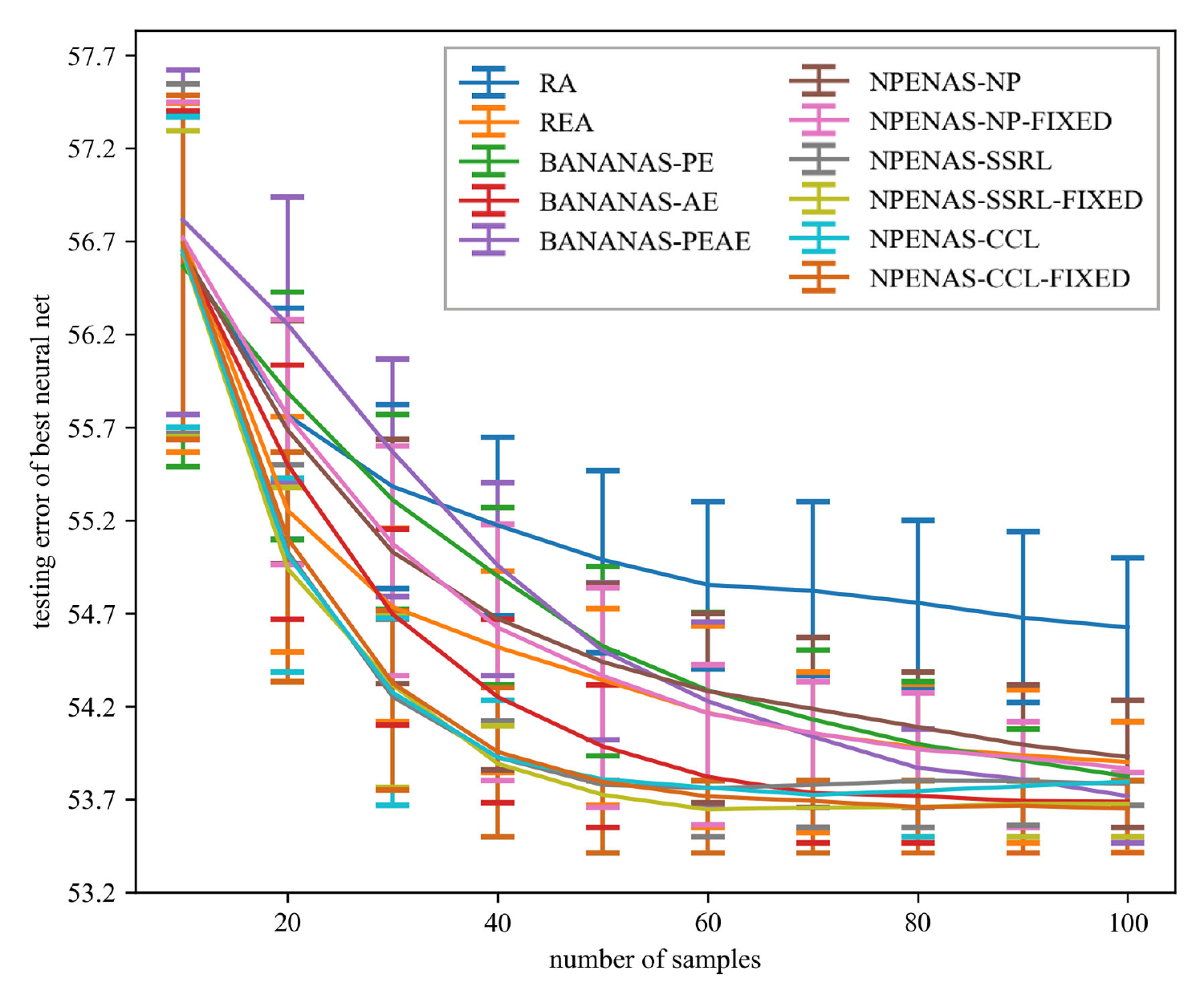}
      \caption{Performance of NAS algorithms on ImageNet-16-120 on NASBench-201.}
      \label{fig_10}
  \end{center}
\end{figure}

We compare above algorithms on the CIFAR-10, CIFAR-100, and ImageNet-16-120 on NASBench-201, and the results are shown in Fig. \ref{fig_8}, Fig. \ref{fig_9} and Fig. \ref{fig_10}, respectively. The quantitative comparison is presented in Table \ref{table_3}. As \textit{arc2vec} \cite{DBLP:journals/corr/abs-2006-06936} does not report queries on this benchmark, we do not compare with it.

As can be seen in Fig. \ref{fig_8}, Fig. \ref{fig_9}, and Fig. \ref{fig_10}, all algorithms can find neural architectures with good performance using a small search budget. As shown in Table \ref{table_3}, NPENAS-SSCCL achieves the best performance on both CIFAR-100 and ImageNet-16-120 on NASBench-201, with nearly the \textit{ORACLE} baseline on CIFAR-100 ($26.5 \%$ vs. $26.49 \%$). On ImageNet-16-120, the performance of NPENAS-SSCCL is equivalent to the \textit{ORACLE} baseline. On CIFAR-10 on NASBench-201, NPENAS-SSRL with the fixed search budget achieves the best performance, which is comparable with the \textit{ORACLE} baseline ($8.92 \%$ vs $8.91\%$).

\begin{table*}[ht!]
  \small
  \begin{center}
  \caption{Performance comparison of NAS algorithms on NASBench-201}
  \label{table_3}
  \begin{threeparttable}[b]
  \begin{tabular}{cccccc}
      \hline
      Methods  & \thead{Search \\ Budget}  & \thead{Test Err (\%) Avg \\ CIFAR-10}  & \thead{Test Err (\%) Avg \\ CIFAR-100} & \thead{Test Err (\%) Avg \\ ImageNet-16-120} \\ \hline
      RA \cite{Li2019RandomSA} &  100  &  9.26 $\pm$ 0.32 & 28.54 $\pm$ 0.87 & 54.62 $\pm$ 0.83 \\
      REA \cite{Real2018RegularizedEF} &  100  & 8.97 $\pm$ 0.22  & 27.16 $\pm$ 0.85 & 53.9 $\pm$ 0.67 \\ 
      BANANAS-PE \cite{White2019BANANASBO} & 100 & 9.05 $\pm$ 0.3 & 27.37 $\pm$ 0.99 & 53.82 $\pm$ 0.64 \\ 
      BANANAS-AE \cite{White2019BANANASBO}  & 100 & 8.96 $\pm$ 0.16 & 26.75 $\pm$ 0.66 & 53.69 $\pm$ 0.38 \\
      BANANAS-PEAE \cite{White2019BANANASBO} & 100 & 8.94 $\pm$ 0.16 & 26.9 $\pm$ 0.74 & 53.7 $\pm$ 0.5 \\
      NPENAS-NP \cite{Chen2019NPENAS} & 100 & 8.95 $\pm$ 0.13 & 26.74 $\pm$ 0.67 & 53.9 $\pm$ 0.62 \\
      NPENAS-NP-FIXED  \cite{Chen2019NPENAS} & 50$^\dagger$ & 8.94 $\pm$ 0.13 & 26.7 $\pm$ 0.51 & 53.87 $\pm$ 0.57 \\ \hline
      NPENAS-SSRL  & 150 & 8.94 $\pm$ 0.1 & 26.51 $\pm$ 0.21 & 53.78 $\pm$ 0.43 \\
      NPENAS-SSRL-FIXED  & 50$^\dagger$ & \textbf{8.92 $\pm$ 0.1} & 26.57 $\pm$ 0.29 & 53.68 $\pm$ 0.4 \\
      NPENAS-SSCCL  & 150 & 8.94 $\pm$ 0.1 & \textbf{26.5 $\pm$ 0.15} & \textbf{53.8 $\pm$ 0.32} \\
      NPENAS-SSCCL-FIXED  & 50$^\dagger$ & 8.94 $\pm$ 0.11 & 26.57 $\pm$ 0.3 & 53.65 $\pm$ 0.36 \\ \hline
  \end{tabular}
  \begin{tablenotes}
    \item[$\dagger$] {\footnotesize The neural predictor is trained with 50 evaluated neural architectures, while the NPENAS algorithm needs 150 evaluate architectures.}
\end{tablenotes}
  \end{threeparttable}
\end{center}
\end{table*}

\section{Conclusion}
We present a new neural architecture encoding scheme, position-aware path-based encoding, to calculate the GED of neural architectures. To enhance the performance of neural predictors, we propose two self-supervised learning methods to pre-train the neural predictors' architecture embedding modules to generate a meaningful representation of neural architectures. Extensive experiments illustrate the superiority of the self-supervised pre-training. When integrating the pre-trained neural predictors with NPENAS, we achieve the state-of-the-art performance on the NASBench-101 and NASBench-201 benchmarks.

The experimental results show that the two self-supervised learning pre-trained neural predictors illustrate totally different behaviors. An in-depth investigation and theoretical analysis are needed to uncover the mechanism that leads the difference, which will help the design of a better self-supervised learning method for NAS. Combining the pre-trained neural predictors to other neural predictor-based NAS to verify the generalize ability of the pre-trained neural predictors is worthy for further study. Extending the interaction of pre-trained neural predictors with NPENAS to other tasks like image segmentation, object detection, and natural language processing is also a meaningful future work.

\appendices
\section{The Operation Position Analyze of Path-based Encoding Scheme} \label{appendices_1}
\begin{figure}[ht!]
  \begin{center}
      \begin{subfigure}{0.23\textwidth}
          \includegraphics[width=0.85\linewidth, height=4.5cm]{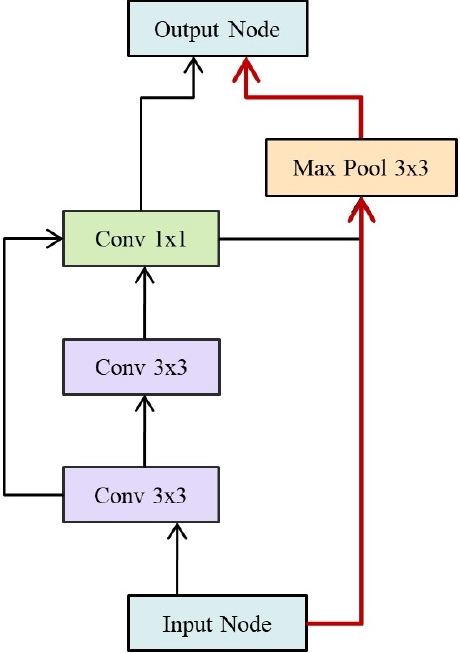}
          \caption{}
          \label{appendix_1_sub_1}
      \end{subfigure}
      \begin{subfigure}{0.23\textwidth}
          \includegraphics[width=0.9\linewidth, height=4.5cm]{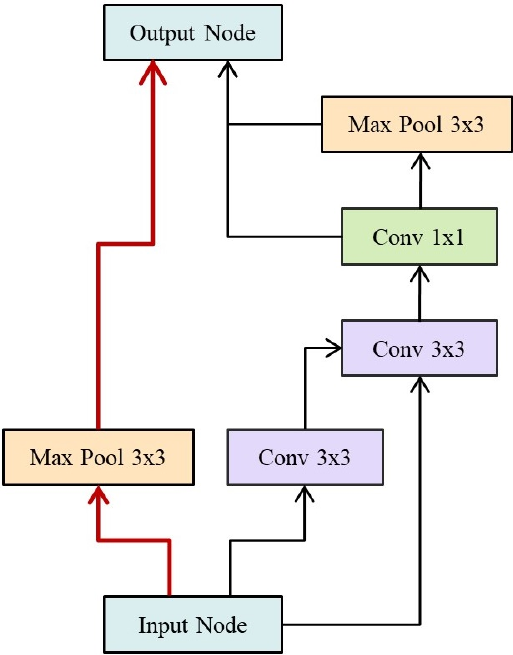}
          \caption{}
          \label{appendix_1_sub_2}
      \end{subfigure}
      \caption{(a) and (b) are two different neural architectures from the NASBench-101 search space.}
      \label{appendix_1}
  \end{center}
\end{figure}

Fig. \ref{appendix_1} shows two neural network architectures from the NASBench-101 search space, and the mean percentage test accuracy of the two neural architectures are $91.2 \%$ (\ref{appendix_1_sub_1}) and $90.4 \% $ (\ref{appendix_1_sub_2}), respectively. 

\begin{figure}[ht!]
  \begin{center}
      \includegraphics[width=0.8\linewidth, height=5.5cm]{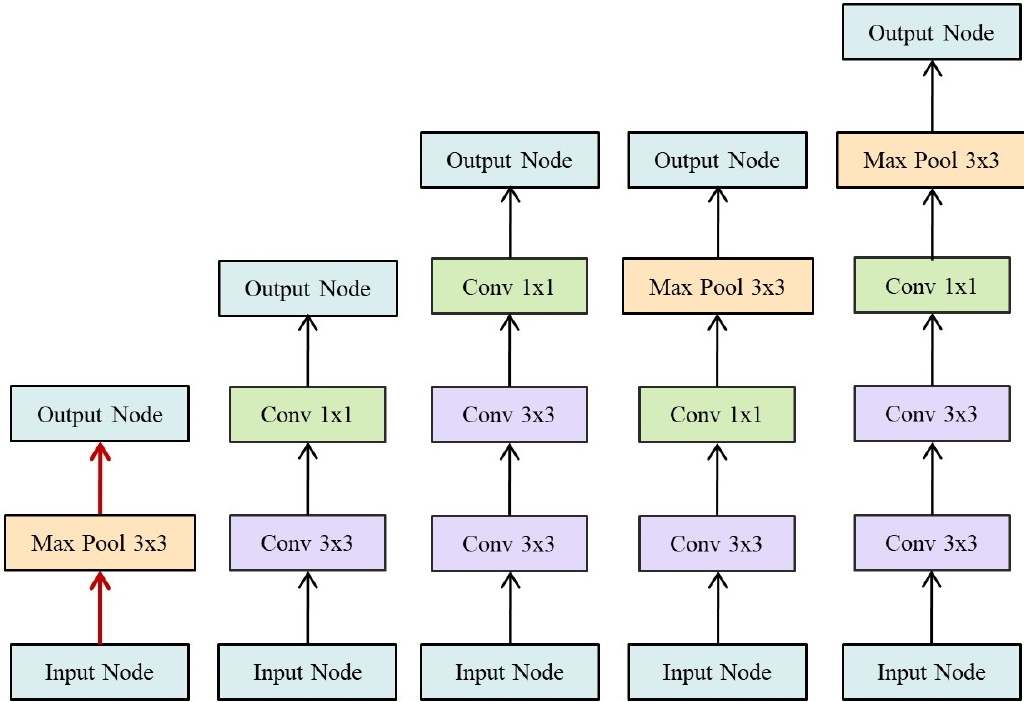}
      \caption{The path-based encoding of the two neural architectures in Fig. \ref{appendix_1}.}
      \label{appendix_3}
  \end{center}
\end{figure}

The two neural architectures in Fig. \ref{appendix_1} have the same path-based encoding, as shown in Fig. \ref{appendix_3}. The red line path in Fig. \ref{appendix_1_sub_1} and Fig. \ref{appendix_1_sub_2} indicates the same input-to-output path that only contains a max-pool 3$\times$3 operation. Although the red line path in the two neural architectures is identical, the position of the max-pool 3x3 operation in the two neural architectures are different, wherein Fig. \ref{appendix_1_sub_1}, the input of the max-pool 3x3 operation is the input node and the convolution 1x1 operation, and in Fig. \ref{appendix_1_sub_2} the input of the max-pool 3x3 operation is the input node. The ignorance of the position of operations in the neural architecture caused the path-based encoding method to map the two different neural architectures in Fig. \ref{appendix_1} into the same encoding vector.

\section{An Illustration of the Central Contrastive Learning} \label{appendices_2}
As illustrated in Fig. \ref{fig_4}, the objective of the central contrastive learning is to aggregate the positive green features to the center vector $\textbf{e}_c$ and push the negative orange features far away from the center.

\begin{figure}[ht!]
  \begin{center}
      \includegraphics[width=0.7\linewidth, height=4.4cm]{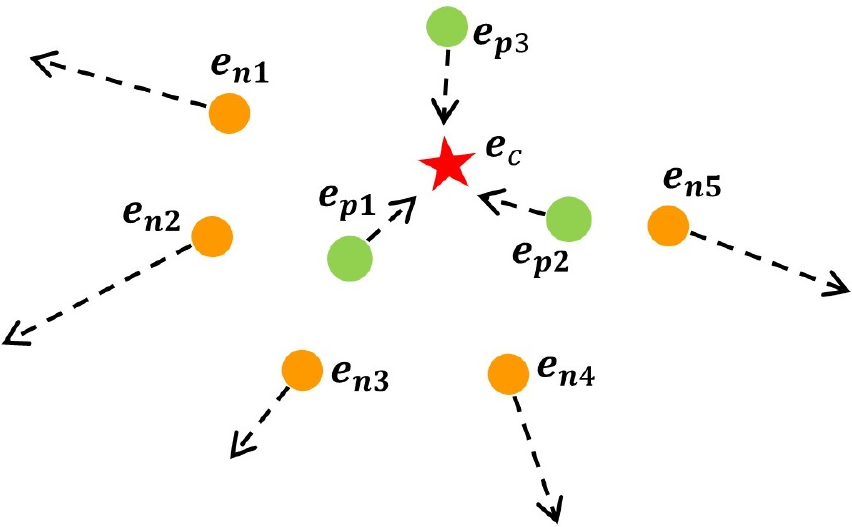}
      \caption{An illustration of the central contrastive learning.}
      \label{appendix_4}
  \end{center}
\end{figure}

The central vector $\textbf{e}_c$ in Fig. \ref{appendix_4} is calculated by average all the green feature vectors and can be formulated as 
\begin{equation}
  \textbf{e}_c = \sum_{i=1}^{3} \textbf{e}_{pi}.
\end{equation}

The contrastive loss corresponding to Fig. \ref{appendix_4} is defined as 
\begin{equation}
  l = \sum_{i=0}^{3}-\log \frac{\exp(sim(\textbf{e}_{pi}, \textbf{e}_c))}{\exp(sim(\textbf{e}_{pi}, \textbf{e}_c)) + \sum_{k=1}^{5}\exp(sim(\textbf{e}_{nk}, \textbf{e}_c))}  .
\end{equation}

\ifCLASSOPTIONcaptionsoff
  \newpage
\fi

\bibliographystyle{IEEEtran}
\bibliography{reference_db}

\end{document}